\PassOptionsToPackage{table,xcdraw}{xcolor}
\PassOptionsToPackage{numbers,sort&compress}{natbib}
\documentclass{article} 
\usepackage[preprint]{neurips_2026}

\usepackage{microtype}
\usepackage{hyperref}
\usepackage{url}
\usepackage{booktabs}
\usepackage{multirow}
\usepackage{float}
\usepackage{placeins}
\usepackage{needspace}
\usepackage{dblfloatfix}
\usepackage{graphicx}
\usepackage{caption}
\usepackage{amsmath}
\usepackage{amssymb}
\usepackage{mathtools}
\usepackage[most]{tcolorbox}
\tcbuselibrary{breakable}

\definecolor{darkblue}{rgb}{0, 0, 0.5}
\hypersetup{colorlinks=true, citecolor=darkblue, linkcolor=darkblue, urlcolor=darkblue}

\newtcolorbox{promptbox}[2][]{
    colback=gray!5,
    colframe=black!60,
    coltitle=white,
    fonttitle=\bfseries\small,
    title={#2},
    breakable,
    enhanced,
    fontupper=\footnotesize\ttfamily,
    lines before break=2,
    boxrule=0.8pt,
    #1
}

\title{AVCap: Reinforcing Audio-Video Joint Caption with Detail-Aware Reward}

\author{Mingyang Wu$^{1}$, Kaituo Feng$^{1}$, Bohao Li, Kaixiong Gong$^{1}$, \\
\bfseries Zihao Yin$^{2}$, Xiangyu Yue$^{1}$\thanks{Corresponding author.} \\
$^{1}$ MMLab, CUHK \\
$^{2}$ Peking University}

\begin{document}

\setcounter{footnote}{1}
\maketitle
\setcounter{footnote}{0}
\raggedbottom

\begin{abstract}
Detailed audio-video joint captioning is essential for multimodal video understanding and generation. However, prior works are constrained by three main limitations: (1) the scarcity of high-quality public datasets with fine-grained audio-visual joint captions; (2) reinforcement-learning methods that rely on coarse reward signals; and (3) the lack of a benchmark and metric for evaluating detailed audiovisual captions at the atomic level. To address these challenges, we propose: (1) \textbf{AVCap-100K}, a high-quality dataset of 100K temporally aligned, detail-rich audio-video captions; (2) \textbf{AVCap}, a model optimized via Detail-Aware GRPO (Da-GRPO) that achieves state-of-the-art performance among open-source models and matches or surpasses proprietary models on several evaluations; and (3) \textbf{AVCap-Bench \& AVCap-Score}, a specialized benchmark and metric for evaluating atomic-level details in audiovisual captions. Our code, models, and datasets are available at \url{https://huggingface.co/collections/Apryle/avcap}.

\end{abstract}

\vspace{-1em}
\section{Introduction}
Recent advancements in Multimodal Large Language Models (MLLMs) have sparked a growing research interest in fine-grained audio-visual captioning \citep{chen2025avocado,wu2025ugc}.
For understanding, detailed captions characterize events with both temporal precision and content richness. This granularity enables models to capture specific attributes and their exact synchronization, facilitating a deep and fine-grained alignment between visual and auditory information \citep{zhou2025daily}.
Simultaneously, for generation, this semantic richness establishes a foundation for advanced synthesis. The dense information provides the necessary priors for high-fidelity generation and supports strict adherence to complex instructions for precise controllability \citep{chen2025vidcapbench,chen2024sharegpt4video}.

However, despite significant advancements in audio-visual joint video captioning \citep{shi2025mavors,ren2024timechat,wang2024tarsier,yuan2025tarsier2}, the field is constrained by three structural deficiencies: (1) the scarcity of high-quality public datasets specifically annotated for fine-grained audio-visual details, as existing resources are typically vision-centric or lack sufficient semantic granularity \citep{chen2024panda,xue2022advancing}; (2) the application of Reinforcement Learning (RL) in audio-visual captioning remains constrained by the granularity of reward signals. Existing approaches primarily rely on event-level checklists rewards \citep{chen2025avocado}, which optimize for narrative coverage but often overlook atomic-level factual precision; and (3) the inadequacy of exsisting benchmarks for evaluating high-granularity understanding, which generally focus on coarse event-level alignment \citep{tang2025video,sungbin2025avhbenchcrossmodalhallucinationbenchmark,geng2025longvale}. Consequently, existing methods struggle to deliver captions that are both factually precise and semantically comprehensive.

To address these challenges, we present a comprehensive framework that redefines the standards for detailed audio-visual captioning.
First, to dismantle the data bottleneck, we curate \textbf{AVCap-100K}, a high-quality dataset of 100K video-caption pairs sourced from audio-visually rich content and annotated with atomic-level density. To mitigate the audiovisual mismatch \citep{xu2025mitigating} inherent in unified pipelines, we implement a rigorous annotation strategy based on explicit unimodal disentanglement, independently extracting dense visual and hierarchical audio priors (ASR, vocal, BGM) before synthesizing them through joint reasoning. Crucially, this pipeline incorporates temporal integration to ensure coherent long-form narratives and enforces a rigorous rating-based filter across audio, visual and joint dimensions to guarantee atomic-level fidelity. 
Methodologically, we propose \textbf{AVCap}, a model optimized via our novel \textbf{Detail-Aware GRPO (Da-GRPO)}. Based on Group Relative Policy Optimization, or GRPO \citep{shao2024deepseekmath}, this approach addresses the limitations of prior RL methods in video captioning, which predominantly rely on coarse, holistic rewards that fail to discern atomic-level hallucinations or omissions. To enforce precise controllability, \textbf{Da-GRPO} reformulates the reward mechanism as a dense informational verification process following a Raise-Answer-Check paradigm. Specifically, a judge model first raises a comprehensive set of atomic-level probes derived from the ground truth, focusing on auditory, visual details and their alignment. The core of this mechanism is quantifying the informational consistency of atomic facts, where the judge evaluates the semantic similarity between the answers deduced from the generated caption and the gold-standard answers derived from the ground truth. This comparison yields a atomic-level, dense, and detail-aware reward signal that strictly measures whether the predicted caption accurately preserves the fine-grained visual and auditory details present in the original video, thereby guiding the policy to minimize hallucinations. 

Finally, we establish \textbf{AVCap-Bench} and the \textbf{AVCap-Score} metric, serving as the evaluation counterpart to our training framework. Extending the atomic precision of Da-GRPO to benchmark, this metric transcends surface-level lexical overlap by employing a QA-based verification protocol. It utilizes a judge model to probe for specific visual, auditory, and joint facts within the generated caption, quantitatively measuring its informational equivalence to the human-verified ground truth, which offers precise standard for evaluating joint audio-visual semantic alignment. 

Extensive empirical evaluations demonstrate that \textbf{AVCap} achieves consistently superior performance across diverse benchmarks. Notably, \textbf{AVCap-7B-SFT} outperforms existing open-source baselines of comparable scale, demonstrating the benefit of \textbf{AVCap-100K}, while Da-GRPO provides further gains at both 7B and 30B scales. Scaling further, \textbf{AVCap-30B} achieves scores of 56.94 on AVCap-Bench, 85.1 on UGC-VideoCap, and 32.7 on Video-SALMONN-2. These metrics indicate that our model delivers comparable or superior performance against proprietary commercial model Gemini-2.5-Pro \citep{comanici2025gemini} on several evaluations.

Our contributions can be summarized as follows:
\begin{itemize}
    \setlength{\itemsep}{3pt}
    \setlength{\parsep}{0pt}
    \setlength{\parskip}{0pt}
    \item We release \textbf{AVCap-100K}, a high quality dataset of 100K video-caption pairs annotated with superior audio-visual density. It serves as a critical resource to resolve the community's scarcity of high quality fine-grained training data.
    
    \item We introduce \textbf{AVCap}, a model trained via our novel Detail-Aware-GRPO (Da-GRPO). AVCap achieves an average score of \textbf{85.1} on UGC-VideoCap and a total score of \textbf{32.7} on Video-SALMONN-2, outperforming existing open-source baselines and remaining competitive with proprietary models including Gemini-2.5-Pro.
    
    \item We propose \textbf{AVCap-Bench} and \textbf{AVCap-Score}, a specialized benchmark and metric designed to rigorously quantify fine-grained informational content, offering the first atomic-level evaluation standard for audiovisual captioning.
\end{itemize}

\section{Related Work}

\begin{figure}[htbp]
    \centering
    \vspace{-1em}
    \includegraphics[trim=160 265 140 190,
    clip,
    width=1\linewidth]{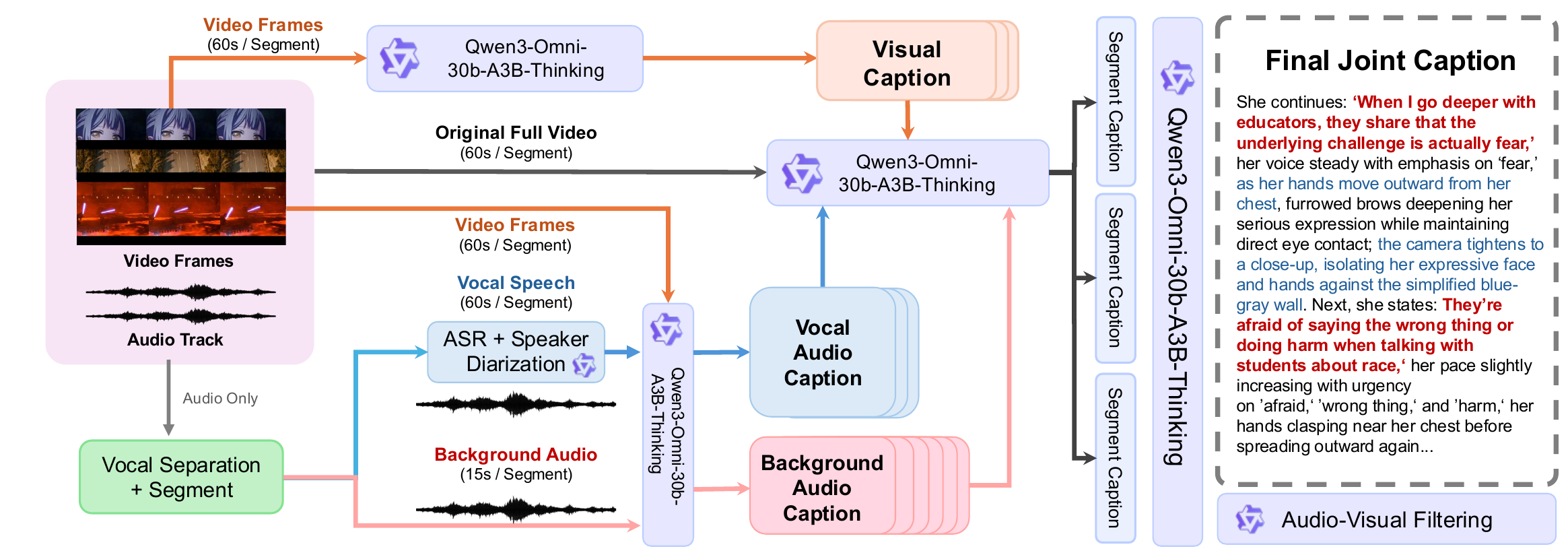}
    \caption{\textbf{Annotation Pipeline of AVCap-100K.} The pipeline first performs \textbf{Dynamic Segmentation}, then extracts unimodal priors through a \textbf{Visual Branch} and a hierarchical \textbf{Audio Branch} with Demucs-based source separation, ASR, Vocal Captioning, and BGM Captioning. These priors are fused in \textbf{Joint Reasoning} to produce the final audio-visual caption, followed by rating-based quality filtering.}
    \label{fig: datapipeline}
    \vspace{-1em}
\end{figure}

\subsection{Large Language Models for Video Captioning} Video Large Language Models (VideoLLMs) have evolved from vision-centric approaches to unified omni-modal architectures \citep{wang-etal-2018-watch}. Early vision-only models \citep{xu2021videoclip,bai2025qwen2,wang2024tarsier,xu2025glavecapgloballocalalignedvideo} combined visual encoders with LLMs to generate dense descriptions, yet they inherently lack auditory context. To address this, Omni-modal models \citep{xu2025qwen2,comanici2025gemini,lu2025omnicaptioner} employ token interleaving to fuse audio-visual signals. While recent state-of-the-art methods have made strides in joint understanding, they face distinct limitations: AVoCaDO \citep{chen2025avocado} emphasizes temporal alignment; Omni-Captioner \citep{ma2025omni} focuses on detailed synthesis but remains confined to supervised paradigms; and UGC-VideoCaptioner \citep{wu2025ugc} introduces RL but utilizes coarse holistic rewards limited to short-form content. Consequently, a gap remains in leveraging fine-grained RL to achieve atomically precise audio-visual captioning.

\subsection{Reinforcement Learning for Video Understanding}
Inspired by DeepSeek-R1 \citep{guo2025deepseek} and Group Relative Policy Optimization (GRPO), Reinforcement Learning (RL) has become central to video understanding. For general tasks, GRPO is widely used to strengthen multimodal reasoning \citep{feng2025video,wang2025adatooler,park2025deepvideo,li2025editthinker,fan2025sophiavl,wu2025reinforcing} and improve temporal grounding \citep{wang2025time,feng2025onethinker}, enhancing long-form video processing. For example, Video-R1 \citep{feng2025video} introduced T-GRPO to enhance temporal-aware video reasoning capabilities, and demonstrated promising performance.
In video captioning, GRPO-based RL has shown promise by boosting visual event recall and temporal synchronization \citep{li2025videochat,meng2025videocap,tang2025video}. Yet these advances largely remain confined to the visual modality only and overlook the semantics carried by audio \citep{shen2024exploring}. Extending RL to audiovisual captioning, prior work \citep{chen2025avocado,wu2025ugc} employs RL to improve the temporal synchronization of audiovisual events, but typically with sparse, event-level rewards. Consequently, leveraging RL to produce dense, detailed descriptions grounded jointly in vision and sound remains largely unexplored.

\section{Method}

\subsection{AVCap-100K Dataset}

Constructing a dataset with high granularity and precise temporal alignment requires careful handling of multi-modal interactions. However, existing pipelines often suffer from incomplete modality separation and lack dynamic segmentation strategies. Consequently, they fail to balance fine-grained audio-visual details effectively and lead to severe information loss in long-form videos. To overcome these obstacles, we design a rigorous annotation pipeline based on a multi-stage, multi-perspective strategy, as illustrated in Figure \ref{fig: datapipeline}. By explicitly decoupling unimodal priors before synthesizing them, our approach ensures the generation of a high-quality, detail-rich dataset. This pipeline consists of four key stages:

\noindent \textbf{Video Segmentation.} 
Rigid fixed-window segmentation often fractures continuous events. To address this, we adopt a dynamic sliding window strategy. Videos are segmented into 60-second units. Crucially, for trailing segments, clips shorter than 10 seconds are merged into the preceding segment to preserve semantic integrity, preventing the generation of fragmented, context-poor captions common in static processing.

\noindent \textbf{Fine-Grained Modal-Wise Extraction.}
To mitigate the cross-modal hallucination where visual models infer sound or vice versa, we employ independent pipelines to extract dense priors for each modality using \textit{Qwen3-Omni-Thinking}:

 \textbf{1. Visual Branch:} To capture purely visual details without auditory bias, we feed the mute video stream (60s) into the model. This isolation forces the model to rely solely on visual cues, generating a comprehensive description free from auditory interference.

 \textbf{2. Hierarchical Audio Branch:} Overlapping speech and background noise severely degrade ASR and captioning accuracy in standard pipelines. To address this, we first robustly decouple the soundtrack into vocal and background tracks using Demucs \citep{rouard2023hybrid}.

\begin{itemize}
    \item \textbf{Vocal Caption:} The clean vocal stream is then processed by Qwen3-Omni-Thinking for high-fidelity ASR, after which the isolated audio, video frames, and transcripts are jointly used to attribute speaker characteristics such as tone, emotion, and gender, yielding a structured Vocal Caption.
    \item \textbf{BGM Captions:} In parallel, to capture temporal dynamics often missed by global summaries, we segment the non-vocal background track into 15-second windows and generate independent BGM Captions, thereby modeling rapid acoustic transitions and environmental shifts at a higher temporal resolution.
\end{itemize}

\noindent \textbf{Audio-Visual Joint Reasoning \& Temporal Integration.}
To mitigate modality dominance common in end-to-end generation, we employ \textit{Qwen3-Omni-Thinking} for joint reasoning. By conditioning on rich unimodal priors (Visual, ASR, Vocal, and BGM captions) alongside raw audiovisual data, the model generates a Joint Caption that explicitly aligns visual events with auditory cues. Additionally, for videos exceeding 60s, a Temporal Integration step synthesizes segment-level outputs into a coherent narrative.

\noindent \textbf{Rating-based Filtering.}
Publicly sourced data inevitably contains noise. To ensure reliability, we implement a rigorous quality control mechanism. The generated joint captions are evaluated by \textit{Qwen3-Omni-Thinking} across three dimensions: Visual Accuracy, Audio Fidelity, and Alignment Precision. Only samples achieving high confidence scores are retained. To quantify the residual noise after this automatic filtering stage, we further conduct a manual verification on 1,000 randomly sampled training examples in Appendix \ref{app:data_validation}.

\begin{figure}[H]
    \centering
    \begin{minipage}[t]{0.49\linewidth}
        \centering
        \makebox[\linewidth][c]{%
            \includegraphics[height=0.16\textheight]{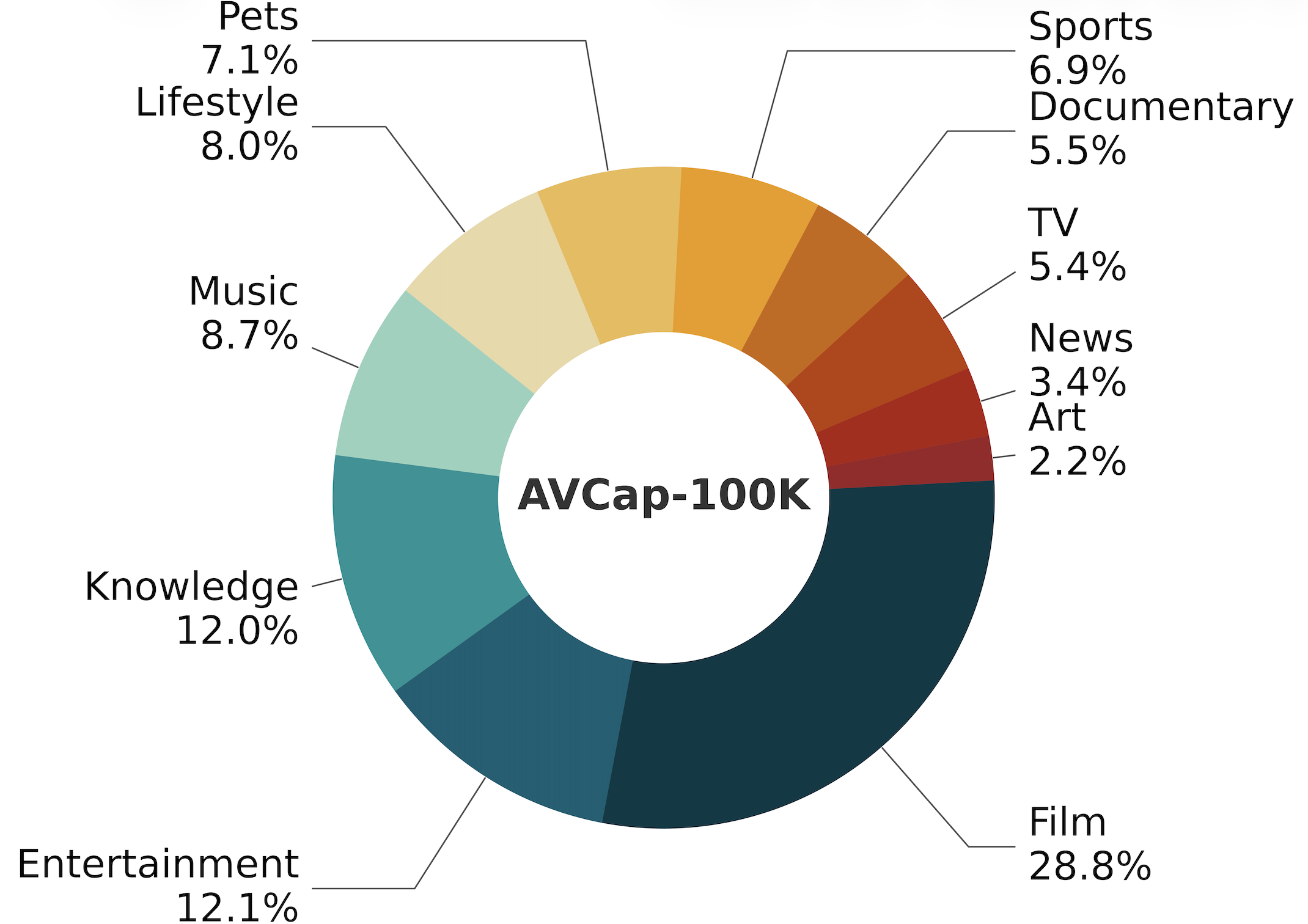}%
        }
        \captionof{figure}{\textbf{Distribution of Video Categories in AVCap-100K.}}
        \label{fig:category_stats}
    \end{minipage}\hfill
    \begin{minipage}[t]{0.49\linewidth}
        \centering
        \makebox[\linewidth][c]{%
            \includegraphics[height=0.16\textheight]{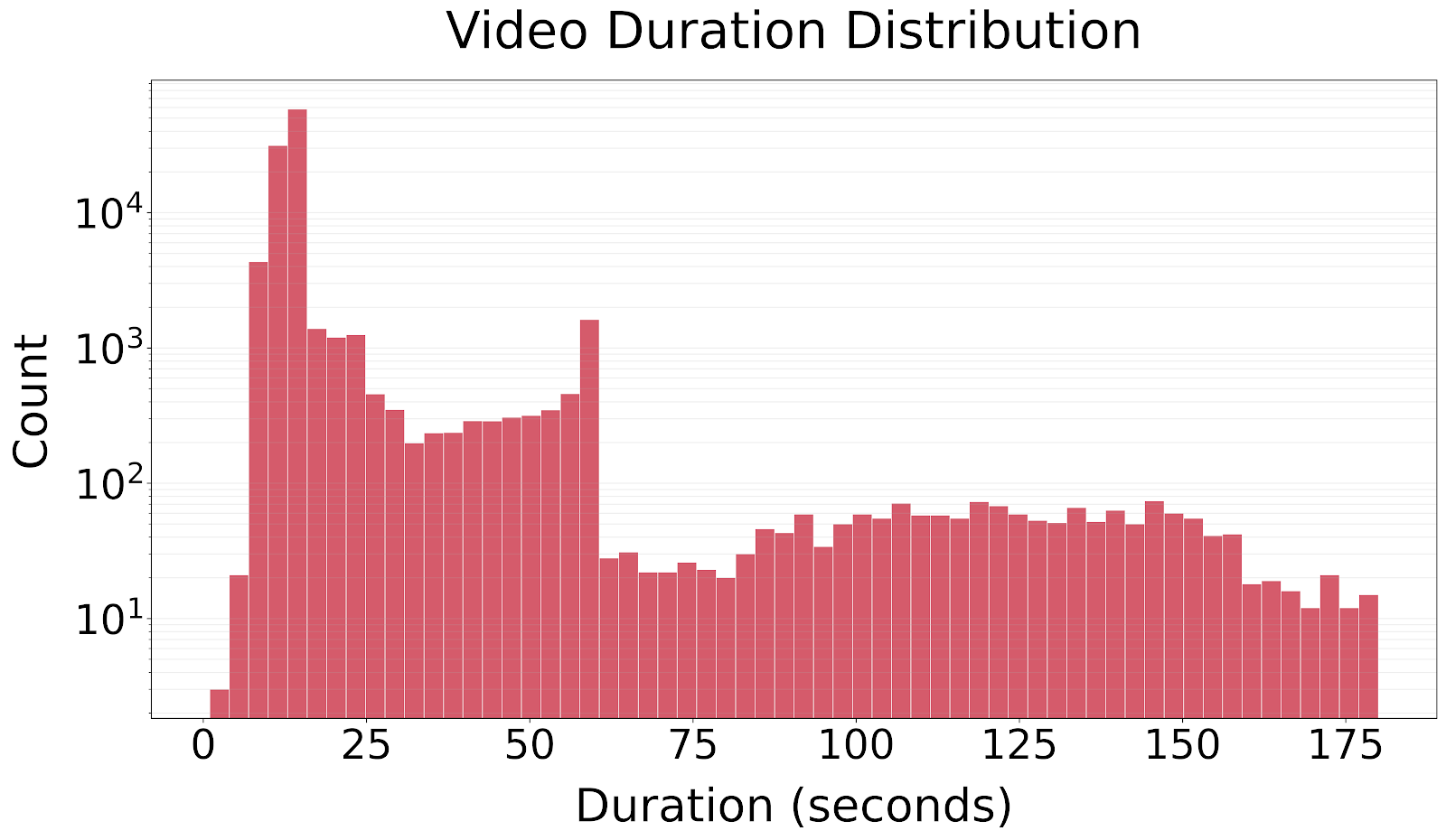}%
        }
        \captionof{figure}{\textbf{Video Duration Distribution of AVCap-100K.}}
        \label{fig:duration_stats}
    \end{minipage}
\end{figure}

\noindent \textbf{Data Curation \& Diversity.} We curate a composite source pool integrating established benchmarks (including AVE \citep{tian2018ave}, VGG-Sound \citep{chen2020vggsound}, Condensed Movies \citep{bain2020condensed}, AVQA \citep{yang2022avqa}, Trailer30K \citep{huang2020movienet}, and MPII-MVAD \citep{rohrbach2015dataset}) and  footage from YouTube. Crucially, prior to processing, we employ \textit{Qwen3-Omni-Instruct} to conduct quality filtering and semantic classification on these source videos. This ensures high audio-visual richness in dataset source video clips. As shown in Figure \ref{fig:category_stats} and \ref{fig:duration_stats}, this curation strategy yields a wide distribution across semantic categories and video durations.

\subsection{Detail-Aware-GRPO (Da-GRPO)}

\noindent \textbf{Da-GRPO Reward Formulation.}
Inspired by \citet{chaiauroracap}, to quantify informational density, we utilize a fact-checking mechanism anchored in the ground truth. During the dataset construction phase, the Judge Model $\mathcal{M}_{\text{judge}}$ first raises $N$ granular questions based on the ground-truth caption $C^{\text{gt}}$. These questions are explicitly categorized into three distinct subsets: Visual ($\mathcal{Q}_v$), Audio ($\mathcal{Q}_a$), and Audio-Visual Joint interactions ($\mathcal{Q}_{av}$). Subsequently, the Judge Model answers these questions using $C^{\text{gt}}$ as the oracle context, yielding canonical answers $a^{\text{gt}}_k$. This results in a predefined probe set $\mathcal{Q} = \mathcal{Q}_v \cup \mathcal{Q}_a \cup \mathcal{Q}_{av}$, comprising the collection of all question-answer pairs $\{ (q_k, a^{\text{gt}}_k) \}_{k=1}^N$.

\begin{figure*}[htbp]
    \vspace{-1em}
    \centering
    \includegraphics[trim=140 340 140 200,
    clip,
    width=1\linewidth]{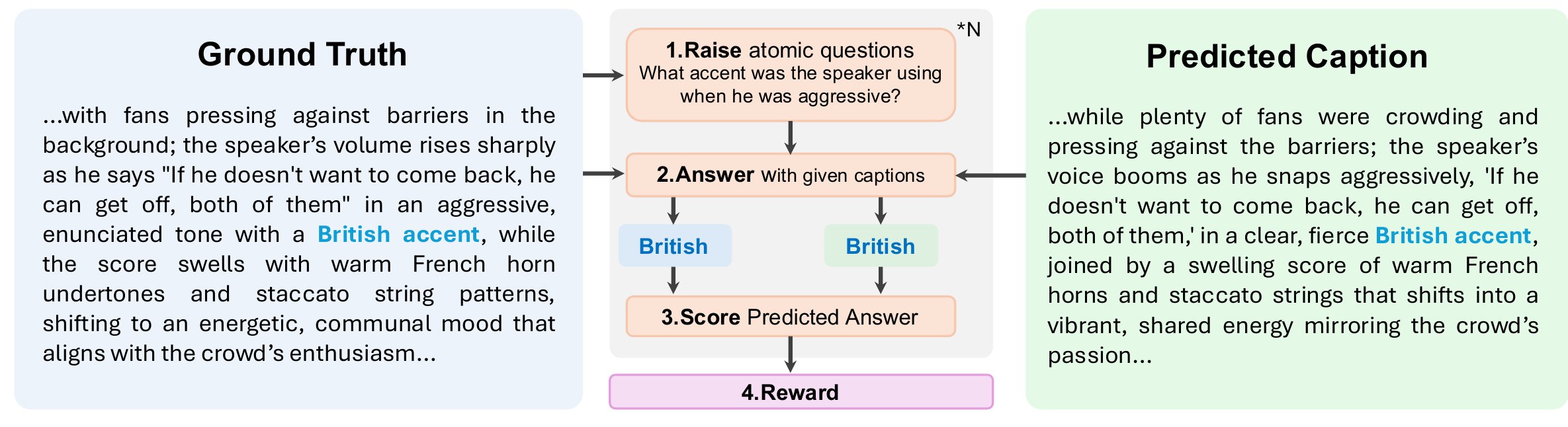}
    \caption{\textbf{Overview of the Da-GRPO Reward calculation.} 
    The workflow follows a \textbf{Raise-Answer-Check} paradigm: (1) \textbf{Raise}: A set of fine-grained audio-visual probes is retrieved from the dataset; (2) \textbf{Answer}: The generated caption acts as the sole context for a Judge Model to answer these probes; (3) \textbf{Score}: The predicted answers are semantically verified against ground-truth answers to compute a dense, atomic-level reward.}
    \label{fig:reward}
    \vspace{-1em}
\end{figure*}

For a candidate generated caption $\hat{C}_i$, the reward computation proceeds through a three-stage process, as illustrated in Figure \ref{fig:reward}:

\begin{enumerate}
    \item[(i)] \textit{Caption-Based Answer Generation.} 
    We treat the generated caption $\hat{C}_i$ as the sole context to answer the probe question $q_k$. The predicted answer $\hat{a}_{i,k}$ is generated as:
    \begin{equation}
        \hat{a}_{i,k} = \mathcal{M}_{\text{judge}}(\cdot \mid \text{Context}=\hat{C}_i, \text{Query}=q_k)
    \end{equation}
    This step tests whether the atomic fact queried by $q_k$ is recoverable from $\hat{C}_i$.

    \item[(ii)] \textit{Semantic Verification.} 
    Next, we evaluate the correctness of $\hat{a}_{i,k}$ against the ground truth $a^{\text{gt}}_k$. The judge assigns a semantic similarity score $s_{i,k} \in [0, S_{\text{max}}]$ (where $S_{\text{max}}=5$):
    \begin{equation}
        s_{i,k} = \text{Score}\left( \hat{a}_{i,k}, a^{\text{gt}}_k ; \mathcal{M}_{\text{judge}} \right)
    \end{equation}

    \Needspace{7\baselineskip}
    \item[(iii)] \textit{Dense Reward Aggregation.} 
    The raw reward $r_i$ for caption $\hat{C}_i$ is the normalized average across the probe set $\mathcal{Q}$, ensuring balanced attention to visual, audio, and joint details:
    \begin{equation}
        r_i = \frac{1}{N \cdot S_{\text{max}}} \sum_{k=1}^{N} s_{i,k}
    \end{equation}
\end{enumerate}

\noindent \textbf{Da-GRPO Optimization Objective.}
With the atomic detail-aware rewards computed, we employ Group Relative Policy Optimization (GRPO) to update the policy. Given a video input $x$, the policy model $\pi_\theta$ samples a group of candidate captions $\{\hat{C}_i\}_{i=1}^G$ from the old policy $\pi_{\theta_{\text{old}}}$. 

The advantage $\hat{A}_i$ can be computed as:
\begin{equation}
    \hat{A}_i = \frac{r_i - \text{mean}(\{r_1, \dots, r_G\})}{\text{std}(\{r_1, \dots, r_G\})}
\end{equation}
Finally, the GRPO objective maximizes this advantage subject to a KL-divergence constraint:
\begin{equation}
\label{eq:grpo}
\mathcal{J}_{\text{Da-GRPO}}(\theta) = \mathbb{E}_{x \sim \mathcal{D}, \{\hat{C}_i\}_{i=1}^G \sim \pi_{\theta_{\text{old}}}(\cdot|x)} 
\left[ \frac{1}{G} \sum_{i=1}^G \left( \frac{\pi_\theta(\hat{C}_i|x)}{\pi_{\theta_{\text{old}}}(\hat{C}_i|x)} \hat{A}_i - \beta \mathbb{D}_{\text{KL}}(\pi_\theta || \pi_{\text{ref}}) \right) \right]
\end{equation}
By explicitly maximizing this objective, the policy $\pi_\theta$ is incentivized to generate descriptions that are not only fluent but also semantically detailed and factually aligned with the video content.

\begin{figure*}[!t]
    \centering
    \includegraphics[trim=120 220 170 250,
    clip,
    width=1\linewidth]{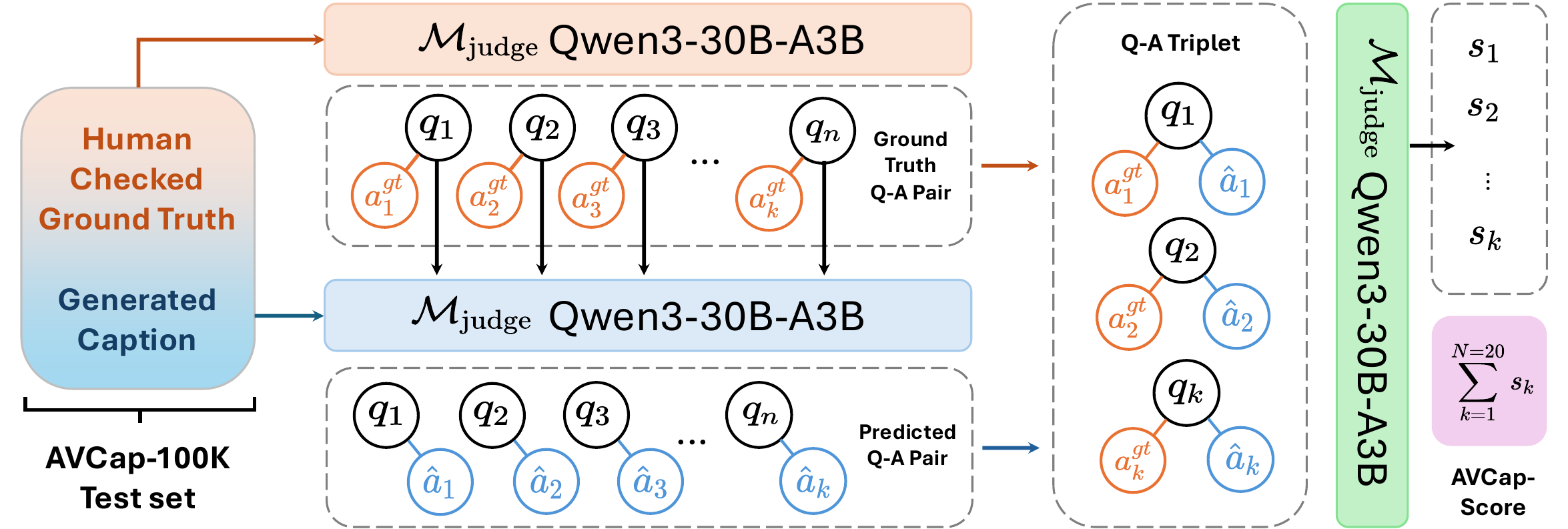}
    \caption{\textbf{Evaluation Pipeline of AVCap-Score.} The metric compares informational density through a QA proxy: (1) the top branch derives 20 atomic QA pairs from the human-checked ground-truth caption; (2) the bottom branch answers the same questions using the generated caption as the only context; and (3) the right branch semantically verifies the predicted answers against the ground truth to produce the final score.}
    \label{fig:avcap_score_pipeline}
\end{figure*}

\subsection{AVCap-Bench and AVCap-Score}

To tackle challenges of \textbf{Modal Disentanglement} and \textbf{Atomic Granularity} while constructing a benchmark for detailed audio-visual captioning, we establish \textbf{AVCap-Bench}. Its 1,000 examples come from a held-out split fixed before training, and each ground-truth caption is manually checked for fine-grained correspondence with the audiovisual signals.

 To quantify performance, we propose \textbf{AVCap-Score}, an automated metric that measures the informational equivalence between the video and the generated caption. Unlike traditional n-gram metrics (e.g., CIDEr) that rely on surface-level lexical overlap, AVCap-Score decomposes the evaluation into specific atomic facts, mirroring the granularity of our training objective, following protocols of multimodal post-training works that directly optimize benchmark-aligned core metrics as rewards \citep{wangtime,xu2025mixed,shen2025vlm}.

\noindent \textbf{Benchmark Construction \& Pipeline.}
We construct the test set $\mathcal{D}_{\text{test}}$ by sampling 1,000 videos from AVCap-100K. As illustrated in Figure \ref{fig:avcap_score_pipeline}, the evaluation proceeds in two branches. Given the human-verified caption $C^{gt}$, the Judge Model (Qwen3-30B-A3B), denoted as $\mathcal{M}_{\text{judge}}$, first raises 20 atomic questions $q_k \sim \mathcal{M}_{\text{judge}}(\text{Question} \mid C^{gt})$ and derives oracle answers \(a^{gt}_k = \mathcal{M}_{\text{judge}}(\text{Answer} \mid q_k, C^{gt})\), yielding the Gold Standard probe set $\mathcal{Q} = \{(q_k, a^{gt}_k)\}_{k=1}^{20}$. For a candidate caption $\hat{C}$, the same judge predicts \(\hat{a}_k = \mathcal{M}_{\text{judge}}(\text{Answer} \mid q_k, \hat{C})\), and then assigns a semantic similarity score \(s_k \in [0, 5]\) by comparing \(\hat{a}_k\) against \(a^{gt}_k\) with the Judge Model. This score directly measures whether the atomic fact queried by $q_k$ is recoverable from the generated caption.

\noindent \textbf{Question Generation Principles and Score Calculation.}
The validity of this pipeline relies on the quality of the probes. We adhere to three principles: the answer must be (1) deterministic, (2) inferable from context, and (3) focused on audiovisual granularity. Accordingly, the 20 questions cover three complementary categories: \textbf{Visual Scores ($\mathcal{Q}_v$)}, which probe object colors, camera movements, OCR text, and spatial relationships; \textbf{Audio Scores ($\mathcal{Q}_a$)}, which probe timbre, pitch, specific instruments, background noise, and speaker identity; and \textbf{Joint Scores ($\mathcal{Q}_{av}$)}, which probe temporal synchronization, causality, and source grounding. To ensure interpretability, we normalize the score of each modality to a 0--100 scale. Let $s_k \in [0, 5]$ denote the semantic similarity score assigned by the Judge Model for the $k$-th question. We define the sub-scores for Visual (Visual-Score), Audio (Audio-Score), and Joint (Joint-Score) capabilities as \( S_{\text{Type}} = \frac{100}{|\mathcal{Q}_{\text{Type}}| \times 5} \sum_{k \in \mathcal{Q}_{\text{Type}}} s_k \), where Type $\in \{v, a, av\}$. Finally, the total \textbf{AVCap-Score} is computed as \( \text{AVCap-Score} = \sum_{k=1}^{|\mathcal{Q}|} s_k \), with \( s_k \in [0, 5] \). We additionally audit 100 generated probe--answer pairs to verify that they target caption-grounded fine-grained facts (Appendix~\ref{app:data_validation}).

\section{Experiment}
\subsection{Experimental Settings}

\noindent \textbf{Implementation Details.}
We use \textbf{Qwen3-Omni-30B-Instruct} (30B-A3B) as the main backbone and train on 32 $\times$ 80GiB GPUs across four nodes. Training proceeds in two stages: \textbf{Stage I} performs 1 epoch of full-parameter SFT on 60K videos from AVCap-100K with a global batch size of 128, a learning rate of 1e-5, cosine decay, and MoE parallelism (Tensor=2, Expert=4, Pipeline=2, Context=2), taking about 5 hours; \textbf{Stage II} initializes from the SFT checkpoint and applies GRPO for 200 steps with LoRA \citep{hu2022lora} (Rank=128, $\alpha$=256), $G=8$, $N=20$, a global batch size of 512, a learning rate of 5e-5, $\beta=0.01$, and vLLM \citep{kwon2023efficient} for efficient rollout, taking about 23 hours. We also train \textbf{AVCap-7B-SFT} from Qwen2.5-Omni-7B and apply Da-GRPO to obtain \textbf{AVCap-7B}; full hyper-parameters and network topologies are provided in Appendix \ref{app:implementation}.

 \begin{table*}[htbp]
\centering
\caption{\textbf{Performance Comparison on General Audiovisual Caption Benchmarks.} We report Hallucination rates (Hall.) and Missing event rates (Miss) on Video-SALMONN-2 \citep{tang2025video}, performance on UGC-VideoCap \citep{wu2025ugc}, and QA-based caption evaluation scores on DailyOmni \citep{zhou2025daily} and WorldSense \citep{hong2025worldsenseevaluatingrealworldomnimodal}.}
\label{tab:main_results}
\resizebox{\linewidth}{!}{%
\small
\renewcommand{\arraystretch}{1.3} 
\setlength{\tabcolsep}{3pt} 

\resizebox{\textwidth}{!}{%
\begin{tabular}{lccccccccccc}
\toprule
\multirow{2}{*}{\textbf{Model}} & \multirow{2}{*}{\textbf{Size}} & \multirow{2}{*}{\textbf{Modality}} & \multicolumn{3}{c}{\textbf{Video-SALMONN-2}} & \multicolumn{4}{c}{\textbf{UGC-VideoCap}} & \multicolumn{2}{c}{\textbf{QA-Based Eval}} \\
\cmidrule(lr){4-6} \cmidrule(lr){7-10} \cmidrule(lr){11-12}
 & & & Miss $\downarrow$ & Hall. $\downarrow$ & Total $\downarrow$ & Audio $\uparrow$ & Visual $\uparrow$ & Detail $\uparrow$ & Avg. $\uparrow$ & DailyOmni $\uparrow$ & WorldSense $\uparrow$ \\
\midrule

\rowcolor{gray!15}\multicolumn{12}{l}{\textit{Proprietary Models}} \\
\midrule
\color{gray}Gemini-2.5-Pro & \color{gray}- & \color{gray}A + V & \color{gray}18.1 & \color{gray}13.3 & \color{gray}31.3 & \color{gray}69.5 & \color{gray}74.7 & \color{gray}73.7 & \color{gray}72.6 & \color{gray}60.2 & \color{gray}33.8 \\
\color{gray}Gemini-2.5-Flash & \color{gray}- & \color{gray}A + V & \color{gray}19.3 & \color{gray}13.9 & \color{gray}33.3 & \color{gray}69.1 & \color{gray}75.8 & \color{gray}74.0 & \color{gray}73.0 & \color{gray}55.3 & \color{gray}31.0 \\

\midrule
\rowcolor{gray!15}\multicolumn{12}{l}{\textit{Visual-Only Baselines}} \\
\midrule
InternVL3.5 & 8B & V & 53.8 & 25.5 & 79.4 & 47.9 & 64.8 & 59.5 & 57.4 & -- & -- \\
Qwen2.5-VL & 7B & V & 40.5 & 17.0 & 57.5 & 46.6 & 69.1 & 62.3 & 59.3 & -- & -- \\

\midrule
\rowcolor{gray!15}\multicolumn{12}{l}{\textit{Open-Source Audio-Visual Models}} \\
\midrule
UGC-VideoCaptioner & 3B & A + V & 31.6 & 17.0 & 48.6 & 61.4 & 58.4 & 57.5 & 59.1 & 17.0 & 11.2 \\
HumanOmniV2 & 7B & A + V & 49.2 & 12.3 & 61.6 & 45.6 & 66.3 & 59.5 & 57.1 & 8.2 & 6.6 \\
ARC-Hunyuan-Video & 7B & A + V & 45.7 & 12.5 & 58.2 & 52.7 & 56.0 & 55.8 & 54.8 & 8.6 & 8.7 \\
Qwen2.5-Omni & 7B & A + V & 41.7 & 15.4 & 57.1 & 46.9 & 66.1 & 60.0 & 57.7 & 13.4 & 8.6 \\
MiniCPM-o-2.6 & 8B & A + V & 42.2 & 14.3 & 56.5 & 38.6 & 68.5 & 57.7 & 54.9 & 9.8 & 7.2 \\
video-SALMONN-2 & 7B & A + V & \underline{21.2} & 17.6 & 38.8 & 61.8 & 71.4 & 68.5 & 67.2 & 29.9 & 18.2 \\
AVoCaDO & 7B & A + V & \textbf{21.1} & 16.2 & 37.3 & 73.0 & 74.6 & 71.8 & 73.2 & 50.1 & 25.7 \\
Qwen3-Omni-Instruct & 30B-A3B & A + V & 32.0 & 13.6 & 45.6 & 67.5 & 74.8 & \underline{72.3} & 71.5 & 17.5 & 12.7 \\
Qwen3-Omni-Captioner & 30B-A3B & A + V & 31.0 & 16.6 & 47.6 & 69.0 & 75.5 & \underline{72.3} & 72.5 & 27.2 & 14.1 \\
\midrule
\rowcolor{blue!5} \textbf{AVCap-7B-SFT (Ours)} & 7B & A + V & 25.5 & \underline{11.3} & \underline{36.8} & \underline{75.6} & \underline{77.9} & 72.1 & \underline{75.2} & \underline{50.5} & \underline{30.7} \\
\rowcolor{blue!5} \textbf{AVCap-30B (Ours)} & 30B-A3B & A + V & 22.4 & \textbf{10.3} & \textbf{32.7} & \textbf{84.1} & \textbf{87.6} & \textbf{83.7} & \textbf{85.1} & \textbf{52.1} & \textbf{34.3} \\
\bottomrule
\end{tabular}%
}}
\end{table*}

\noindent \textbf{Benchmarks \& Evaluation Protocols.}
We evaluate our model on representative benchmarks that cover both high-level audiovisual semantics and fine-grained details. First, we use the Video-SALMONN-2 testset \citep{tang2025video}, following AVoCaDo \citep{chen2025avocado}, employing GPT-4.1 as the judge to assess general audio-visual understanding. Second, we test on UGCVideoCap \citep{wu2025ugc}, a benchmark for wild user-generated content, where we follow the official protocol using GPT-4o as the evaluator. We also report results on DailyOmni and WorldSense under the QA-based caption evaluation protocol introduced by AVoCaDO \citep{chen2025avocado}. Finally, we report results of our proposed AVCap-Score. Following \citet{huang2024vbench, zheng2025vbench}, all scores are computed with the open-source Qwen3-30B-A3B-Instruct judge to ensure sustainability and reproducibility .

\begin{table*}[!t]
\centering
\caption{\textbf{Fine-grained Evaluation on AVCap-Score Benchmark.} We report the accuracy of atomic fact verification across Visual, Audio, and Joint dimensions, where all scores are computed using Qwen3-30B-A3B-Instruct as the judge.}
\label{avcapscore_results}
\resizebox{\linewidth}{!}{%
\small
\renewcommand{\arraystretch}{1}
\setlength{\tabcolsep}{3pt}

\resizebox{\linewidth}{!}{%
\begin{tabular}{lccccccc}
\toprule
\textbf{Model} & \textbf{Size} & \textbf{Modality} & \textbf{Visual-Score} & \textbf{Audio-Score} & \textbf{Joint-Score} & \textbf{AVCap-Score} \\
\midrule

\rowcolor{gray!15}\multicolumn{7}{l}{\textit{Proprietary Models}} \\
\midrule
 \color{gray}Gemini-2.5-Pro & \color{gray}{--} & \color{gray}{A + V} & \color{gray}{49.70} & \color{gray}{49.70} & \color{gray}{51.53} & \color{gray}{50.62} \\
 \color{gray}Gemini-2.5-Flash & \color{gray}{--} & \color{gray}{A + V} & \color{gray}{51.11} & \color{gray}{47.73} & \color{gray}{51.32} & \color{gray}{50.37} \\

\midrule

\rowcolor{gray!15}\multicolumn{7}{l}{\textit{Open-source Baselines}} \\
\midrule
Qwen2.5-Omni & 7B & A + V & 38.19 & 26.17 & 36.11 & 34.14 \\
HumanOmniV2 & 7B & A + V & 36.96 & 31.74 & 38.56 & 36.45 \\
UGC-VideoCaptioner & 3B & A + V & 38.88 & 35.31 & 39.61 & 38.35 \\
AVoCaDO & 7B & A + V & 47.48 & 46.34 & 45.18 & 46.05 \\
Qwen3-Omni-Instruct & 30B-A3B & A + V & \underline{48.59} & 43.49 & 44.72 & 45.38 \\
Qwen3-Omni-Captioner & 30B-A3B & A + V & 48.22 & 45.94 & 46.70 & 46.89 \\
\midrule

\rowcolor{blue!5} \textbf{AVCap-7B-SFT (Ours)} & 7B & A + V & 47.67 & \underline{50.62} & \underline{50.37} & \underline{49.76} \\
\rowcolor{blue!5} \textbf{AVCap-30B (Ours)} & 30B-A3B & A + V & \textbf{58.10} & \textbf{57.64} & \textbf{56.01} & \textbf{56.94} \\
\bottomrule
\end{tabular}%
}}
    \vspace{-1em}

\end{table*}

\noindent \textbf{Baselines.}
Following the protocol in AVoCaDO \citep{chen2025avocado}, we benchmark against a comprehensive suite of models on Video-SALMONN-2 and UGCVideoCap. We include leading open-source AV models including Qwen2.5-Omni \citep{xu2025qwen2}, AVoCaDO \citep{chen2025avocado}, HumanOmniV2 \citep{yang2025humanomniv2}, ARC-Hunyuan-Video \citep{ge2025arc}, MiniCPM-o-2.6 \citep{openbmb2025minicpm}, Qwen3-Omni-Instruct and Qwen3-Omni-Captioner. We also include commercial models Gemini-2.5 series \citep{comanici2025gemini}, and strong vision-only baselines Qwen2.5-VL \citep{bai2025qwen2}, InternVL3.5 \citep{wang2025internvl3} to isolate audio contributions. For DailyOmni and WorldSense, we evaluate under the same QA-based protocol as AVoCaDO, while the remaining baseline results are quoted directly from AVoCaDO. For our fine-grained \textbf{AVCap-Score}, we prioritize relevant Omni models and specialized competitors, specifically Gemini series, Qwen2.5-Omni, HumanOmniV2, UGC-VideoCaptioner \citep{wu2025ugc}, AVoCaDO \citep{chen2025avocado}, and the Qwen3-Omni family to demonstrate the efficacy and scalable potential of our training pipeline on larger scale models.

\subsection{Experiment Results}

\noindent \textbf{Results on Existing Audio-Visual Caption Benchmarks.}
As presented in Table~\ref{tab:main_results}, our proposed method demonstrates consistent superiority across model scales. On Video-SALMONN-2, \textbf{AVCap-7B-SFT} marginally outperforms the previous open-source SOTA, while \textbf{AVCap-30B} further improves the hallucination rate to \textbf{10.3}, remaining competitive with proprietary systems. On UGC-VideoCap, \textbf{AVCap-7B-SFT} surpasses all baselines, and \textbf{AVCap-30B} reaches an average score of \textbf{85.1}. On DailyOmni \citep{zhou2025daily} and WorldSense \citep{hong2025worldsenseevaluatingrealworldomnimodal}, AVCap-7B-SFT scores \underline{50.5}/\underline{30.7}, while AVCap-30B reaches \textbf{52.1}/\textbf{34.3}, surpassing AVoCaDO and Gemini-2.5-Pro on WorldSense.

\subsection{Ablation Studies of Da-GRPO}

\begin{figure*}[htbp]
    \centering
    \includegraphics[width=1\linewidth]{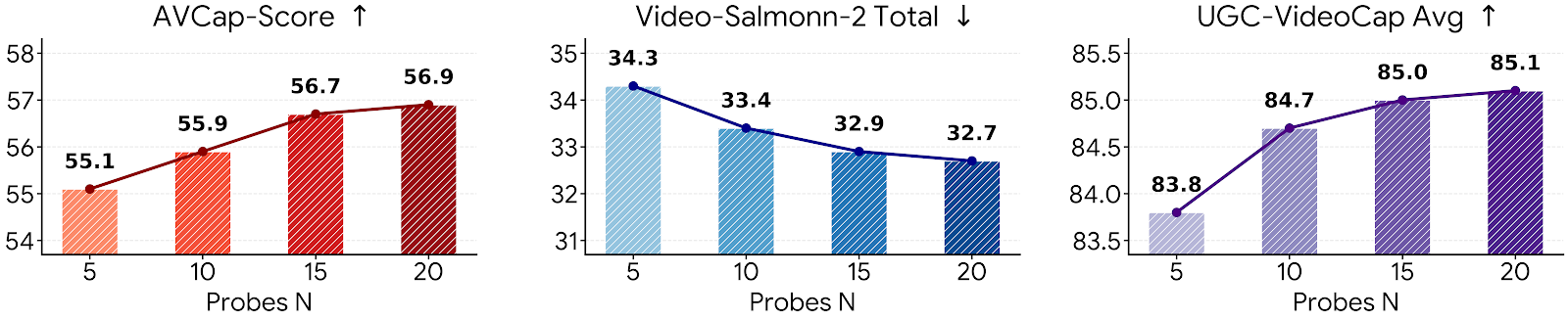}
    \vspace{-0.1in}
    \caption{\textbf{Impact of Probe Question Quantity ($N$).}}
    \label{fig:ablation_q}
    \vspace{-1em}
\end{figure*}

\noindent \textbf{Impact of Probe Question Quantity ($N$).}
We analyze the sensitivity to probe count \( N \) in Figure \ref{fig:ablation_q}. Performance positively correlates with reward density, yet marginal gains diminish as \( N \) approaches 20. This saturation aligns with the intrinsic information capacity of our \(\sim\)15s video clips, a duration also to effectively serve downstream video generation models. While also representing a practical tradeoff between resources and reward effectiveness, \( N=20 \) serves as the empirical optimal, ensuring comprehensive coverage of atomic facts without inducing redundancy.

\noindent \textbf{Impact of Reward Granularity.} To isolate the contribution of our proposed reward mechanism, we conduct ablation studies on the \textbf{AVCap-30B} backbone, comparing \textbf{Da-GRPO} against three representative baselines ranging from standard supervision to coarse-grained reinforcement learning: \textbf{SFT-Only}, which trains the model solely with cross-entropy loss and serves as a lower bound; \textbf{w/ N-gram Reward}, a traditional approach similar to SCST \citep{rennie2017self} that uses BLEU-4 \citep{papineni2002bleu} as a rule-based reward for surface-level lexical overlap; and \textbf{w/ Holistic Semantic Reward}, which directly prompts the Judge Model to assign a single scalar score (1--5) based on the overall semantic similarity between the generated caption and the ground truth. Results in Table~\ref{tab:attribution} separates the SFT and Da-GRPO gains across the dense 7B and MoE 30B-A3B backbones, in Table~\ref{tab:ablation_reward} show that atomic-level reward signals consistently outperform coarse-grained alternatives, confirming the importance of \textbf{Da-GRPO} for precise audio-visual grounding. 

\vspace{-0.5em}
\begin{table}[H]
\centering
\caption{\textbf{Controlled Attribution of AVCap-100K and Da-GRPO.} We report aggregate scores under identical benchmark protocols.}
\label{tab:attribution}
\small
\setlength{\tabcolsep}{5pt}
\resizebox{\linewidth}{!}{%
\begin{tabular}{lcccccc}
\toprule
\textbf{Model} & \textbf{Backbone} & \textbf{AVCap-100K} & \textbf{Da-GRPO} & \textbf{VS2 Total $\downarrow$} & \textbf{UGC Avg. $\uparrow$} & \textbf{AVCap-Score $\uparrow$} \\
\midrule
Qwen2.5-Omni & Qwen2.5-Omni-7B & $\times$ & $\times$ & 57.1 & 57.7 & 34.14 \\
AVoCaDO & Qwen2.5-Omni-7B & $\times$ & $\times$ & 37.3 & 73.2 & 46.05 \\
\midrule
\rowcolor{blue!5} AVCap-7B-SFT & Qwen2.5-Omni-7B & $\checkmark$ & $\times$ & 36.8 & 75.2 & 49.76 \\
\rowcolor{blue!5} \textbf{AVCap-7B} & Qwen2.5-Omni-7B & $\checkmark$ & $\checkmark$ & \textbf{35.9} & \textbf{78.0} & \textbf{51.59} \\
\rowcolor{blue!5} AVCap-30B-SFT & Qwen3-Omni-30B-A3B & $\checkmark$ & $\times$ & 34.3 & 83.5 & 53.03 \\
\rowcolor{blue!5} \textbf{AVCap-30B} & Qwen3-Omni-30B-A3B & $\checkmark$ & $\checkmark$ & \textbf{32.7} & \textbf{85.1} & \textbf{56.94} \\
\bottomrule
\end{tabular}
}
\end{table}
\vspace{-0.5em}

\FloatBarrier
\begin{table*}[htbp]
\centering
\caption{\textbf{Ablation Study of Different Reward Mechanisms.}}
\label{tab:ablation_reward}
\small
\renewcommand{\arraystretch}{1.3}
\setlength{\tabcolsep}{3pt}
\resizebox{\linewidth}{!}{%
\begin{tabular}{l ccc cccc cccc}
\toprule
\multirow{2}{*}{\textbf{Method}} & \multicolumn{3}{c}{\textbf{Video-SALMONN-2}} & \multicolumn{4}{c}{\textbf{UGC-VideoCap}} & \multicolumn{4}{c}{\textbf{AVCap-Score}} \\
\cmidrule(lr){2-4} \cmidrule(lr){5-8} \cmidrule(lr){9-12}
 & Miss $\downarrow$ & Hall. $\downarrow$ & Total $\downarrow$ & Audio $\uparrow$ & Visual $\uparrow$ & Detail $\uparrow$ & Avg. $\uparrow$ & Visual $\uparrow$ & Audio $\uparrow$ & Joint $\uparrow$ & Total $\uparrow$ \\
\midrule
AVCap-30B-SFT & 23.9 & 10.4 & 34.3 & 82.3 & 86.2 & 82.1 & 83.5 & 55.72 & 54.05 & 51.18 & 53.03 \\
w/N-gram Reward & 23.5 & 10.4 & 33.9 & 82.2 & 85.6 & 80.7 & 82.8 & 56.02 & 53.84 & 53.64 & 54.28 \\
w/Holistic Reward & 22.9 & 11.6 & 34.5 & 82.5 & 85.8 & 82.6 & 83.6 & 56.89 & 53.92 & 51.97 & 53.68 \\
\midrule
\rowcolor{blue!5} \textbf{w/Da-GRPO} & \textbf{22.4} & \textbf{10.3} & \textbf{32.7} & \textbf{84.1} & \textbf{87.6} & \textbf{83.7} & \textbf{85.1} & \textbf{58.10} & \textbf{57.64} & \textbf{56.01} & \textbf{56.94} \\
\bottomrule
\end{tabular}%
}
\end{table*}

\vspace{-0.3cm}
\section{Conclusion}
In this paper, we present a holistic framework for fine-grained audio-visual captioning. First, we construct \textbf{AVCap-100K}, a high-quality dataset that empowers a 7B model to surpass existing baselines. Second, we propose the \textbf{AVCap Model}, which utilizes \textbf{Da-GRPO} to incentivize factually grounded descriptions. Its consistent gains on a dense 7B backbone and a substantially different MoE 30B-A3B backbone demonstrate effectiveness across model scales and architectures. Finally, we establish \textbf{AVCap-Bench} and \textbf{AVCap-Score} to verify atomic facts across Visual, Audio, and Joint dimensions.

\bibliography{example_paper}
\bibliographystyle{unsrtnat}

\newpage
\appendix
\onecolumn
\raggedbottom
\setcounter{table}{0}
\renewcommand{\thetable}{A\arabic{table}}
\setcounter{figure}{0}
\renewcommand{\thefigure}{A\arabic{figure}}
\section{Implementation Details}
\label{app:implementation}
\subsection{Training Data Construction}
Following the standard instruction-tuning paradigm, we format our video-captioning data into a conversation template. Each training sample consists of a system instruction defining the model's persona, a user query containing the \texttt{<video>} token and the specific task prompt, and the target assistant response. The structure is detailed below:

\begin{tcolorbox}[colback=gray!5, colframe=black!60, title=Chat Template for Audio-Visual Captioning, fontupper=\small\ttfamily]
\{  \\
  "id": "sample\_id",   \\
  "videos": ["path/to/video\_clip.mp4"],    \\
  "messages": [ \\
    \{  \\
      "role": "system", \\
      "content": "You are Qwen, a virtual human developed by the Qwen Team, Alibaba Group, capable of perceiving auditory and visual inputs, as well as generating text and speech."    \\
    \}, \\
    \{  \\
      "role": "user",   \\
      "content": "<video> Please provide a thorough description of all the content in the video, including every detail. As you describe, ensure that you also cover as much information from the audio as possible, and be mindful of the synchronization between the audio and video as you do so."   \\
    \}, \\
    \{  \\
      "role": "assistant",  \\
      "content": "[Target Ground Truth Caption...]" \\
    \} ] \\ 
\}  \\
\end{tcolorbox}
\subsection{Training Hyperparameters (SFT)}
\label{app:sft_hyperparams}

We perform full-parameter supervised fine-tuning on the Qwen3-Omni-30B-A3B-Instruct backbone using a cluster of 4 nodes, each equipped with 8 $\times$ 80GiB GPUs (Total 32 GPUs). The training framework is built upon Pytorch-DDP, Megatron-LM and SWIFT. 
To ensure reproducibility, we detail the complete configuration in Table \ref{tab:sft_hyperparams}, covering optimization strategies, distributed parallelism setups, and multi-modal data resolution constraints.
\begin{table}[H]
\centering
\caption{\textbf{Detailed Hyperparameters for SFT Stage.} The configuration includes optimization settings, hybrid parallelism strategies for the 30B MoE architecture, and specific constraints for multi-modal data processing.}
\label{tab:sft_hyperparams}
\begin{tabular}{l|c}
\toprule
\textbf{Configuration Group} & \textbf{Value} \\
\midrule
\multicolumn{2}{c}{\textit{Optimization \& Training Dynamics}} \\
\midrule
Global Batch Size & 128 \\
Micro Batch Size & 2 \\
Learning Rate & 1e-5 \\
Min Learning Rate & 1e-6 \\
LR Scheduler & Cosine with Warmup \\
Warmup Ratio & 0.05 \\
Optimizer & AdamW \\
Precision & bfloat16 \\
Max Sequence Length & 32,768 \\
Epochs & 1 \\
Weight Decay & 0.1 \\
Gradient Clipping & 1.0 \\
\midrule
\multicolumn{2}{c}{\textit{Architecture, MoE \& Parallelism}} \\
\midrule
Parallelism Strategy & TP=2, PP=2, EP=4, CP=2 \\
Sequence Parallel & True \\
Distributed Optimizer & True \\
MoE Aux Loss Coeff & 1e-3 \\
MoE Capacity Factor & 2.0 \\
MoE Router & Top-K (K=2) \\
Trainable Parameters & Full LLM (Vision Encoder Frozen) \\
\midrule
\multicolumn{2}{c}{\textit{Multi-modal Input \& Resolution Constraints}} \\
\midrule
Max Video Frames (\texttt{FPS\_MAX\_FRAMES}) & 64 \\
Max Video Tokens (\texttt{VIDEO\_MAX\_TOKEN\_NUM}) & 2,048 \\
Max Image Tokens (\texttt{IMAGE\_MAX\_TOKEN\_NUM}) & 4,096 \\
Max Total Pixels (\texttt{VIDEO\_TOTAL\_PIXELS}) & 20,070,400 (approx. 20M) \\
Single Image Max Pixels & 20,070,400 \\
Video Max Pixels per Frame & 802,816 \\
Audio-Visual Interleaving & Enabled (\texttt{USE\_AUDIO\_IN\_VIDEO}) \\
\bottomrule
\end{tabular}
\end{table}

\subsection{Training Hyperparameters (Da-GRPO)}

  \paragraph{Training topology.}
  We adopt a four-node layout: two training nodes (\textbf{Master-0} and \textbf{Worker-2}) running distributed Megatron GRPO, one rollout node
  (\textbf{Worker-0}) serving vLLM for online sampling, and one reward node (\textbf{Worker-1}) serving the reward model. Each node uses 8 GPUs. Only the
  two training nodes participate in distributed optimization (total 16 ranks).

  \begin{table}[H]
  \centering
  \caption{\textbf{Training Node Hyperparameters (Da-GRPO).}}
  \label{tab:dagrpo_train_hyperparams}
  \begin{tabular}{l|c}
  \toprule
  \textbf{Configuration Group} & \textbf{Value} \\
  \midrule
  \multicolumn{2}{c}{\textit{Optimization \& Training Dynamics}} \\
  \midrule
  Global Batch Size & 512 \\
  Micro Batch Size & 4 \\
  Learning Rate & 5e-5 \\
  Precision & bfloat16 \\
  Epochs & 1 \\
  Max Sequence Length & 16,384 \\
  Max Completion Length & 1,024 \\
  Temperature & 1.2 \\
  Train Type & LoRA \\
  LoRA Rank / Alpha & 128 / 256 \\
  \midrule
  \multicolumn{2}{c}{\textit{GRPO / Reward Configuration}} \\
  \midrule
  Reward Function & Da-GRPO \\
  Steps per Generation & 1 \\
  Num Generations & 8 \\
  KL Beta & 0.01 \\
  Importance Sampling & sequence-level \\
  Epsilon / Epsilon High & 3e-4 / 4e-4 \\
  Dynamic Sampling & False \\
  Overlong Filter & True \\
  Loss Type & GRPO \\
  \midrule
  \multicolumn{2}{c}{\textit{Architecture \& Parallelism}} \\
  \midrule
  Parallelism Strategy & TP=2, PP=2, EP=2, CP=2 \\
  Sequence Parallel & True \\
  Recompute Granularity & selective \\
  Offload Model / Optimizer & True / True \\
  Padding Free & True \\
  \midrule
  \multicolumn{2}{c}{\textit{Multi-modal Constraints}} \\
  \midrule
  Max Total Pixels (\texttt{MAX\_PIXELS}) & 1,003,520 \\
  Max Video Pixels (\texttt{VIDEO\_MAX\_PIXELS}) & 67,584 \\
  Max Video Frames (\texttt{FPS\_MAX\_FRAMES}) & 16 \\
  Max Image Tokens (\texttt{IMAGE\_MAX\_TOKEN\_NUM}) & 1,024 \\
  Max Video Tokens (\texttt{VIDEO\_MAX\_TOKEN\_NUM}) & 384 \\
  Audio-Visual Interleaving & Enabled (\texttt{USE\_AUDIO\_IN\_VIDEO}) \\
  \midrule
  \multicolumn{2}{c}{\textit{Data}} \\
  \midrule
  Training Dataset & \textbf{AVCap-100K} \\
  \bottomrule
  \end{tabular}
  \end{table}

  \begin{table}[H]
  \centering
  \caption{\textbf{Rollout Node (vLLM) Hyperparameters.}}
  \label{tab:dagrpo_rollout_hyperparams}
  \begin{tabular}{l|c}
  \toprule
  \textbf{Configuration} & \textbf{Value} \\
  \midrule
  Tensor Parallel Size & 8 \\
  GPU Memory Utilization & 0.85 \\
  Max Model Length & 16,384 \\
  Max Num Seqs & 256 \\
  \midrule
  Max Total Pixels (\texttt{MAX\_PIXELS}) & 1,003,520 \\
  Max Video Pixels (\texttt{VIDEO\_MAX\_PIXELS}) & 50,176 \\
  Max Video Frames (\texttt{FPS\_MAX\_FRAMES}) & 12 \\
  Max Image Tokens (\texttt{IMAGE\_MAX\_TOKEN\_NUM}) & 1,024 \\
  Max Video Tokens (\texttt{VIDEO\_MAX\_TOKEN\_NUM}) & 256 \\
  Audio-Visual Interleaving & Enabled (\texttt{USE\_AUDIO\_IN\_VIDEO}) \\
  \bottomrule
  \end{tabular}
  \end{table}

  \begin{table}[H]
  \centering
  \caption{\textbf{Reward Node (vLLM) Hyperparameters.}}
  \label{tab:dagrpo_reward_hyperparams}
  \begin{tabular}{l|c}
  \toprule
  \textbf{Configuration} & \textbf{Value} \\
  \midrule
  Reward Model & Qwen3-14B \\ 
  Thinking & Disabled \\
  Tensor Parallel Size & 8 \\
  GPU Memory Utilization & 0.90 \\
  Max Model Length & 16,384 \\
  Max Num Seqs & 64 \\
  Max Num Batched Tokens & 32,768 \\
  Served Model Name & reward-model \\
  \bottomrule
  \end{tabular}
  \end{table}

\subsection{Da-GRPO Training Efficiency}
\label{app:efficiency}

Table~\ref{tab:dagrpo_efficiency} reports the measured runtime breakdown of the 200-step Da-GRPO stage. The complete stage takes approximately 23 hours.

\begin{table}[H]
\centering
\caption{\textbf{Da-GRPO Runtime Breakdown.}}
\label{tab:dagrpo_efficiency}
\begin{tabular}{lc}
\toprule
\textbf{Measurement} & \textbf{Value} \\
\midrule
Number of optimization steps & 200 \\
Average rollout time per step & 178 s \\
Reward-model inference per step & 55.4 s \\
Average full-step time & 424.6 s \\
Reward-computation share & 13.06\% \\
Total wall-clock time & $\sim$23 h \\
Configured vLLM GPU memory utilization (rollout) & 0.85 \\
Configured vLLM GPU memory utilization (reward) & 0.90 \\
\bottomrule
\end{tabular}
\end{table}

 \subsection{Benchmark Implementation Details}

  \paragraph{Pipeline overview.}
  We follow a three-step QA pipeline: (1) caption generation with the tested \textbf{OmniModel}, (2) QA answer generation from captions using a text LLM,
  and (3) automatic grading with the same text LLM. Each step writes JSON outputs and supports resuming.

  \paragraph{Model settings.}
  
  \textbf{OmniModel (Step 1)} is the target model under evaluation; we report only inference settings.
  
  \textbf{Text LLM (Steps 2 \& 3)} defaults to \textit{Qwen3-30B-A3B} for answering and grading.

  \paragraph{vLLM inference configuration.}
  Tensor parallelism is set to \emph{auto} for both OmniModel and the text LLM.
  GPU memory utilization is 0.7 for OmniModel and 0.8 for the text LLM.
  Default maximum model length is 32{,}768; maximum concurrent sequences for OmniModel is 64.
  Audio-visual interleaving is enabled for captioning.

  \paragraph{Decoding parameters.}
  \textbf{Captioning (Step 1)}: temperature 0, max tokens 4096.
  \textbf{QA answering (Step 2)}: temperature 0, max tokens 100.
  \textbf{Grading (Step 3)}: temperature 0, max tokens 5.

  \paragraph{Prompts.}
  All prompts are fixed and shared across models/baselines. The QA answering and grading prompts are used as the system instruction in a chat-style
  template.

  \textbf{QA answering prompt (system instruction).}
  You are a "Grounded Caption Analyst." Your task is to provide a single concise,
  factually correct answer to the provided question based on the video caption.

  Answering Rules:
  
  1. Source of Truth: Base the answer solely on the caption. Use the exact terminology
     found in the text (e.g., if the caption says "shatters", use "shatters", not "breaks").
  
  2. Conciseness: Provide a descriptive phrase or sentence fragment (under 20 words).
  
  3. Directness: Do NOT use filler words like "The answer is..." or
     "According to the caption...". Start directly with the answer content.
  
  4. Format: Escape internal double quotes with a backslash (\textbackslash") if necessary.

  Output Requirement:
  
  Return only the raw answer text string. Do not use quotes around the answer unless they are part of the content.

  \textbf{QA answering chat template.}
  System: \{system instruction above\}
  User: Caption: \{caption\}
  User: Question: \{question\}
  Assistant: (answer only)

  \textbf{Grading prompt (system instruction).}
  
  You are an expert grader evaluating answers about video content.

  Grading Task:
  
  Compare the predicted answer with the ground truth answer and assign a score from 0 to 5.

  Grading Scale:
  
  5: Perfect match or semantically equivalent
  
  4: Mostly correct with minor differences
  
  3: Partially correct, captures main idea
  
  2: Somewhat related but missing key information
  
  1: Incorrect but shows some understanding
  
  0: Completely incorrect or irrelevant

  Output Requirement:
  Provide ONLY the numeric score (0, 1, 2, 3, 4, or 5).
  Do not include any explanation or additional text.

  \textbf{Grading chat template.}
  System: \{system instruction above\}
  User: Question: \{question\}
  User: Ground Truth Answer: \{ground\_truth\_answer\}
  User: Predicted Answer: \{predicted\_answer\}
  Assistant: Score: (single number)

\newpage
\section{AVCap-100K Dataset Details}
\label{app:dataset}

\subsection{Data Source Composition}
\label{app:data_source}

\begin{figure}[H]
    \centering
    \includegraphics[width=0.8\linewidth]{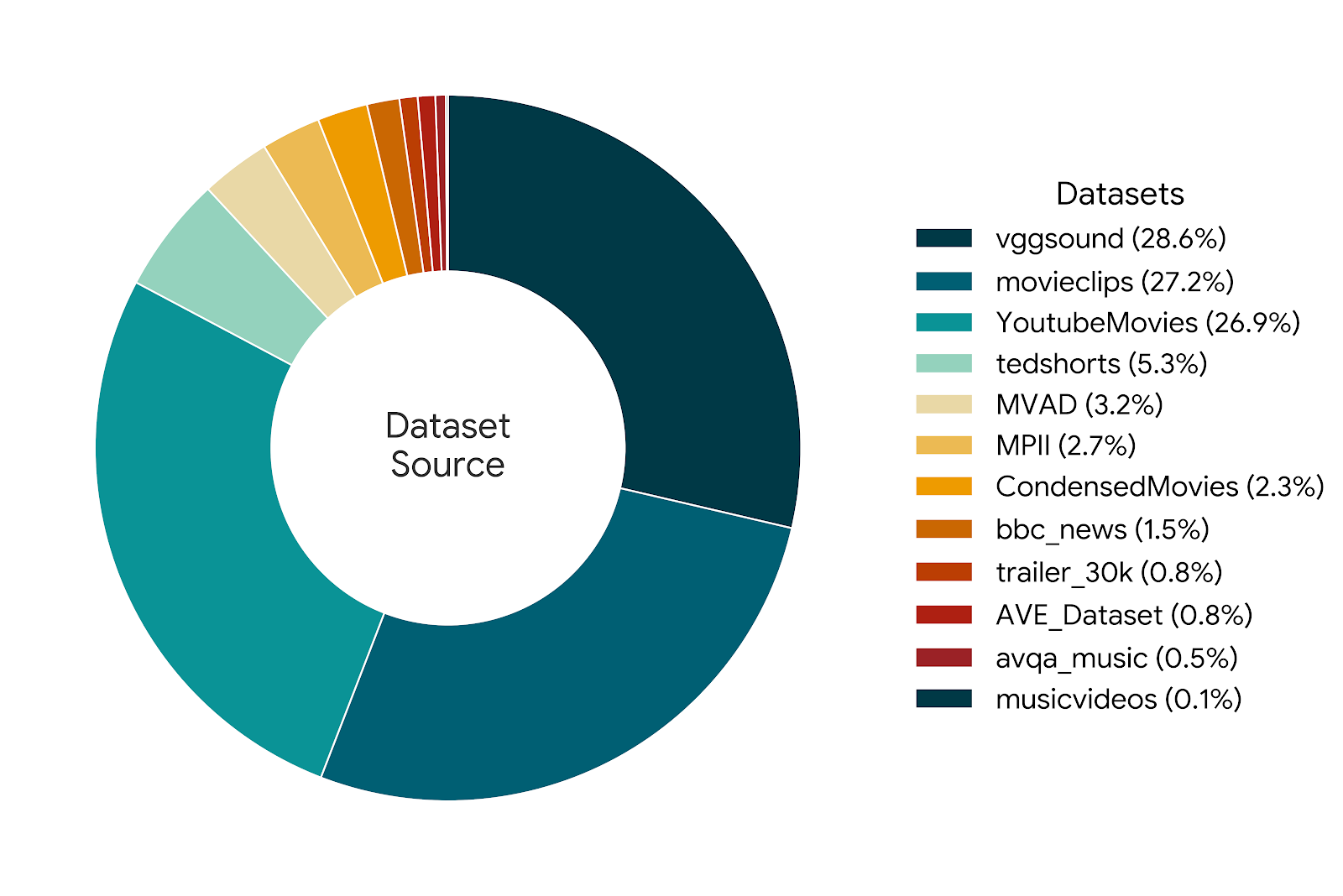} 
    \caption{\textbf{Source Distribution of the AVCap-100K Dataset.} The dataset is composed of 12 diverse sources, with VGGSound, MovieClips, and YTDL serving as the primary foundations.}
    \label{fig:dataset_source}
\end{figure}

To ensure a broad coverage of audio-visual scenarios, we aggregate videos from 12 distinct open-source repositories. Figure \ref{fig:dataset_source} illustrates the detailed distribution of these sources. The dataset is primarily anchored by three large-scale collections: \textbf{VGGSound}, \textbf{MovieClips}, and a  \textbf{YoutubeMovies}, which collectively account for over 82\% of the total data. The remaining portion consists of specialized subsets spanning diverse domains, including news broadcasts (BBC News), educational talks (TED Shorts), and cinematic descriptions (MVAD, Condensed Movies), ensuring the model's robustness across different video styles.

\subsection{Caption Length Distribution}
\label{app:caption_dist}

\begin{figure}[H]
    \centering
    \includegraphics[width=0.8\linewidth]{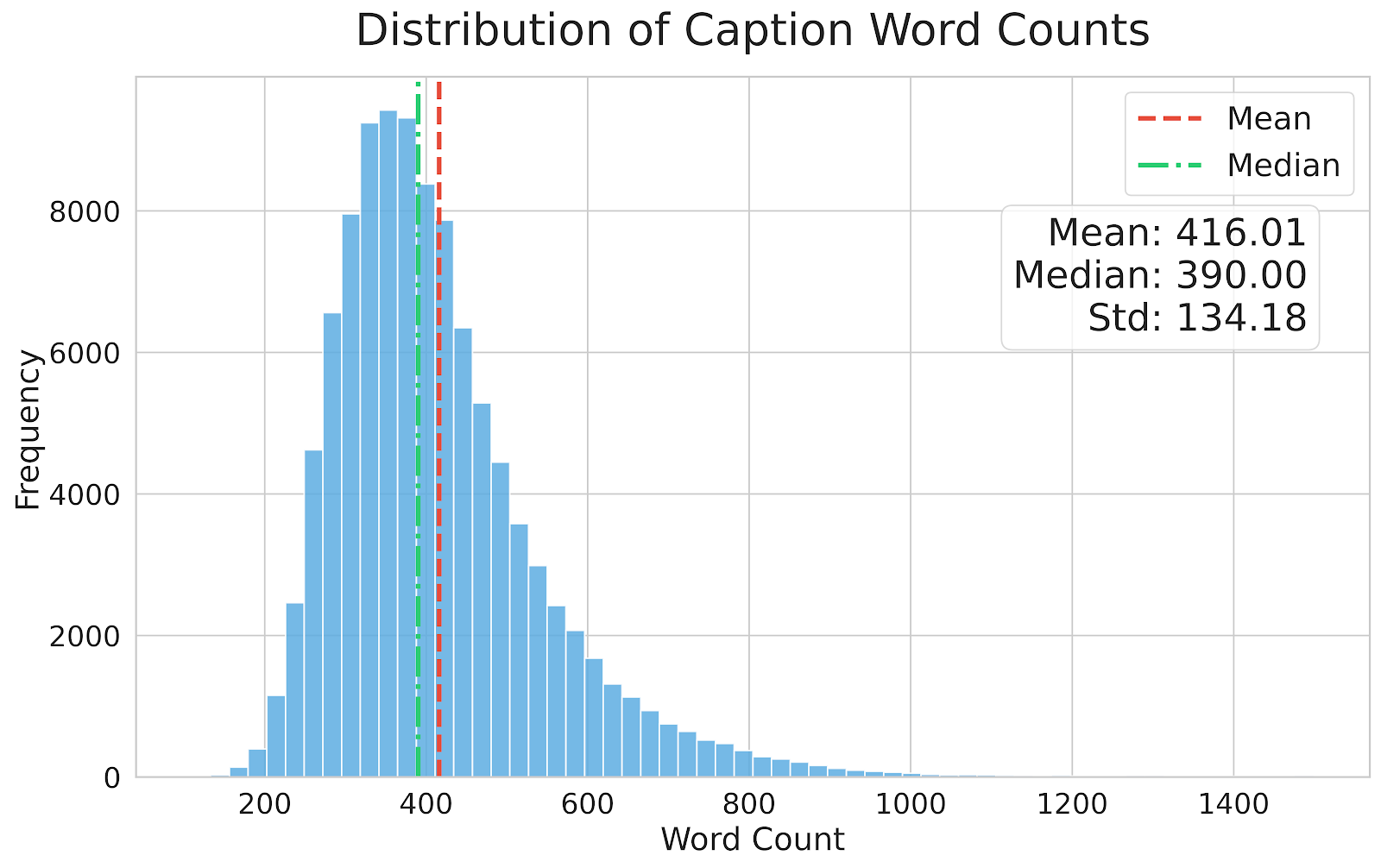} 
    \caption{\textbf{Distribution of Caption Word Counts.} The dataset features dense, long-form descriptions (Mean $\approx$ 416 words), providing rich supervision for fine-grained learning.}
    \label{fig:caption_dist}
\end{figure}

We analyze the lexical density of AVCap-100K to verify its richness. As shown in Figure \ref{fig:caption_dist}, the captions exhibit a robust long-form distribution, with a \textbf{mean length of 416.01 words} and a \textbf{median of 390 words}. The distribution is unimodal and right-skewed, with a standard deviation of 134.18. This high token density is crucial for our objective, as it ensures that the training data contains sufficient capacity to articulate atomic audio-visual details, rather than merely stating high-level events.

\subsection{Validation Audits}
\label{app:data_validation}

We manually audit the data pipeline, generated probes, and caption outputs using the protocols summarized in Table~\ref{tab:validation_audits}.

\begin{table}[H]
\centering
\caption{\textbf{Manual Validation Audits.}}
\label{tab:validation_audits}
\small
\begin{tabular}{lcl}
\toprule
\textbf{Audit} & \textbf{Sample Size} & \textbf{Result} \\
\midrule
ASR transcription & 50 audio samples & WER 4.7\% \\
Judge answers & 100 QA pairs & Error rate 1\% \\
Training captions & 1,000 video--caption pairs & Hallucination rate $<4\%$ \\
Generated probes & 100 probe--answer pairs & No shallow or irrelevant collapse observed \\
Caption verbosity & 100 generated captions & 2\% marked overly verbose \\
\bottomrule
\end{tabular}
\end{table}

For the training-caption audit, a sample is marked as hallucinated if its caption contains an unsupported visual fact, unsupported audio event, or incorrect audio-visual association. For the probe audit, annotators check whether each question targets a fine-grained fact grounded in the reference caption.

\subsection{AVCap-Bench Annotation Protocol}
\label{app:bench_annotation}

The 1,000 AVCap-Bench videos come from a held-out split established before model training and do not overlap with the training set. A team of five annotators reviews each video with audio and checks visual facts, audio facts, and audio-visual temporal alignment. Captions with unsupported facts, missing salient events, or incorrect cross-modal associations are revised or removed, and no sample with an unresolved factual conflict is retained.

\subsection{Human Evaluation}
\label{app:human_evaluation}

We conduct a human evaluation on 100 samples comparing AVCap-30B with AVoCaDO. Five independent annotators score each generated caption for agreement with the ground-truth caption on a 10-point scale. We average the scores across annotators and samples and rescale them to 100 points.

\begin{table}[H]
\centering
\caption{\textbf{Human Evaluation on 100 Samples.}}
\label{tab:human_evaluation}
\begin{tabular}{lcc}
\toprule
\textbf{Model} & \textbf{Human Score $\uparrow$} & \textbf{AVCap-Score $\uparrow$} \\
\midrule
AVoCaDO & 77.20 & 46.05 \\
\rowcolor{blue!5} \textbf{AVCap-30B} & \textbf{84.34} & \textbf{56.94} \\
\bottomrule
\end{tabular}
\end{table}

\subsection{Data Curation Prompts}
\label{app:prompts}

We leverage a suite of specialized prompts to process multi-modal information. To facilitate reproducibility, we provide the full text of the System and User prompts used in our pipeline below. All prompts are presented in \texttt{verbatim} style with automatic line wrapping.

\subsubsection{Visual Analysis Module}

\begin{promptbox}{Visual Captioning - System Prompt}
You are a world-class AI visual analyst and cinematic expert. Your sole purpose is to generate exceptionally detailed, high-granularity, and lengthy visual descriptions of video content in English. You must deconstruct the video into its fundamental visual components with the precision of a seasoned cinematographer and the descriptive power of a novelist. When analyzing a video, you must meticulously describe the following seven core aspects, weaving them together into a coherent, flowing narrative. Do not simply list them as bullet points; integrate them into your prose. 1. Main Subject \& Scene: * Entities: Describe all people (estimating age, gender, attire, hairstyle, facial expressions, posture), animals (species, actions, appearance), and key objects in the foreground and mid-ground. * Details: Note intricate details like fabric textures, logos on clothing, reflections in surfaces, or wear and tear on objects. * Hierarchy: Clearly distinguish between the primary subject of focus and secondary elements. 2. Background \& Environment: * Setting: Is it indoors or outdoors? A specific room (e.g., a cluttered workshop, a minimalist kitchen)? A natural landscape (e.g., a dense forest at dawn, a sun-drenched beach)? An urban environment (e.g., a bustling city street at night, a desolate alleyway)? * Details: Describe architectural styles, furniture, flora and fauna, weather conditions, and the time of day. 3. Action \& Kinetics: * Primary Actions: Detail the main actions of the subjects (e.g., running, talking, dancing, driving). * Micro-actions: Do not neglect subtle movements. Describe a character's shifting gaze, a slight frown, the twitch of a finger, or the rustling of leaves in the wind. * Sequence: Describe actions chronologically as they occur in the clip. 4. Style \& Genre: * Aesthetic: Identify the overall visual style. Is it cinematic, documentary-style, a home video, a corporate video, animation (2D, 3D), a music video, etc.? * Genre Indicators: Based on the visuals, infer the potential genre (e.g., horror, comedy, drama, sci-fi). Mention specific elements that support this, like a shaky camera for horror or vibrant colors for a comedy. 5. Cinematography \& Camera Work: * Movement: Explicitly name the camera movements. Is it static? Is it a slow pan to the right, a tilt upwards, a dolly zoom, a crane shot, a handheld shaky shot, or a smooth Steadicam follow? * Perspective: Describe the camera's point of view. Is it an aerial shot, a top-down (bird's-eye) view, a low-angle shot making the subject look powerful, or a high-angle shot making them look vulnerable? 6. Composition \& Framing: * Shot Type: Identify the framing. Is it an Extreme Wide Shot (EWS), a Wide Shot (WS), a Full Shot (FS), a Medium Shot (MS), a Close-Up (CU), or an Extreme Close-Up (ECU)? * Compositional Rules: Mention the use of compositional techniques like the Rule of Thirds, leading lines, symmetry, depth of field (is the background blurry or sharp?), and framing within a frame. 7. Atmosphere, Lighting \& Color: * Lighting Style: Describe the lighting. Is it high-key (bright, few shadows) or low-key (dark, high contrast)? Is the light hard or soft/diffused? Identify light sources (natural sunlight, a single lamp, neon signs). * Color Palette: Analyze the color grading. Is the scene dominated by a warm (yellows, oranges) or cool (blues, greens) color palette? Is it monochromatic, saturated, or desaturated? * Mood: Conclude by describing the overall mood or feeling created by the combination of light and color (e.g., nostalgic, tense, serene, energetic). Your final output must be a long, descriptive paragraph or a series of paragraphs, written in sophisticated English prose. Your goal is to paint a complete picture in the user's mind, so they can visualize the scene perfectly without ever seeing the video. Do not make your output too long to read.
\end{promptbox}

\begin{promptbox}{Visual Captioning - User Prompt}
Please provide a high-granularity visual caption for the video I have provided. Following the detailed analytical framework, generate a long and comprehensive description in English that covers the Main Subject, Background, Action, Style, Camera Work, Composition, Atmosphere and any other relevant aspects. Describe everything you see. There is no need to mention the audio or sound, as this is a silent video. Your description should be so vivid and detailed that someone reading it can perfectly visualize the entire scene and actions without ever seeing the video.Return the caption content ONLY, no other replies required. Do not make the output too long.

\end{promptbox}

\begin{promptbox}{Visual Filtering - System Prompt}
You are an expert visual content analysis and description model. Your task is to evaluate the accuracy and quality of a given caption based on an accompanying image or video. The caption may contain both visual and audio descriptions. You must **only** focus on the visual-related parts of the caption and ignore any audio information.
Your evaluation must be based on the following four key criteria. For each criterion, you need to provide a detailed analysis and a score from 0 to 5, where 5 is the highest quality.
no "strengths" or "issues" should be mentioned
Your final output **must be a single JSON object** with the following structure: No other response is required.

{
  "evaluation": {
    "visual\_hallucinations": {
      "score": [integer from 0-5],
    },
    "visual\_omissions": {
      "score": [integer from 0-5],
    },
    "visual\_inaccuracies": {
      "score": [integer from 0-5],
    },
    "visual\_ granularity\_and\_detail": {
      "score": [integer from 0-5],
    }
  }
}\end{promptbox}

\begin{promptbox}{Visual Filtering - User Prompt}
please output as plain text insted of markdown, no "strengths" or "issues" should be mentioned. 
\end{promptbox}

\subsubsection{Audio Analysis Module}

\begin{promptbox}{Vocal Captioning - System Prompt}
You are a sophisticated multi-modal AI assistant specializing in high-granularity analysis of human speech in video. You must generate a single, detailed descriptive paragraph synthesizing all analytical components: the verified verbatim transcription, speaker diarization, an in-depth vocal profile (prosody, emotion, demographics), and a description of all synchronized non-verbal cues. The ASR transcrpit is provided, use it as a reference. Return the caption content ONLY, no other replies required. 
\end{promptbox}

\begin{promptbox}{Vocal Captioning - User Prompt}
Analyze the provided video and its reference ASR transcript in strict chronological order. Your task is to produce a highly granular, time-stamped log of all spoken utterances. For each utterance, generate a detailed descriptive paragraph that integrates the corrected word-for-word transcription, the speaker's identifier, a comprehensive vocal profile (perceived demographics, accent, prosody, emotional tone), and a meticulous description of their concurrent non-verbal cues(facial expressions, gestures) and other important visual infomation. The output should be a sequential series of these analytical paragraphs. Do not use speaker A, B, etc from ASR Transcript. Identify speakers by their actual names if mentioned, or according to their roles in the video. If there is no obvious vocal information in the video, point out.
\end{promptbox}

\begin{promptbox}{BGM Captioning - System Prompt}
You are an expert AI audio-visual analyst tasked with creating a meticulous, high-granularity, and chronological log. Your process is to sequentially analyze the video, identify distinct temporal segments based on significant changes in the audio or on-screen action, and for each segment, generate a time-stamped, descriptive paragraph. Each paragraph must offer an in-depth analysis of the audio (music genre/style, mood, instrumentation, soundscape, SFX) and seamlessly integrate this with a description of the corresponding visual context. Your final output should be a narrative sequence of these detailed paragraphs, flowing in chronological order. Do not make the caption too long to read, but we do need fine-grained details.
\end{promptbox}

\begin{promptbox}{BGM Captioning - User Prompt}
Analyze this video's audio-visual track in chronological order, breaking it down into key segments based on changes in the soundtrack or significant on-screen events. For each segment, provide a highly granular and descriptive paragraph. This paragraph should focus primarily on the audio elements (genre, style, mood, instrumentation, soundscape, and any distinct SFX) but must connect them to the simultaneous visual context (Mainly focus on detailed BGM, but not visual information). The goal is a detailed, sequential narrative of the video's sound design and its interplay with the visuals. If there is no obvious BGM/SFX information in the video, point out. Return the caption content ONLY, no other replies required. Controlling length while preserving details.
\end{promptbox}

\begin{promptbox}{Audio Filtering - System Prompt}
You are an expert audio-to-text transcription and analysis model. Your task is to evaluate the accuracy and quality of a given caption based on an accompanying audio file. The caption may contain both visual and audio descriptions. You must **only** focus on the audio-related parts of the caption and ignore any visual information.

Your evaluation must be based on the following four key criteria. For each criterion, you need to provide a detailed analysis:

1.  **Hallucinations:** Identify if there are any audio events or sounds mentioned in the caption that are **not present** in the original audio.
2.  **Omissions:** Identify if there are any significant audio events or sounds present in the original audio that are **missing** from the caption's audio description.
3.  **Inaccuracies:** Identify if any audio events or sounds in the caption are **incorrectly described** compared to the original audio. This includes misidentifications or wrong descriptions of sounds.
4.  **Granularity \& Detail:** Evaluate the level of detail in the audio description. This includes the nuanced description of human speech (e.g., tone, emotion, speaker's state), the richness of background sound effects, and the accuracy of word-level transcriptions for dialogue. We recommend a relatively positive scoring

Your final output **must be a single line JSON object** with the following structure:

{
  "evaluation": {
    "hallucinations": {
      "score": [integer from 0-5],
    },
    "omissions": {
      "score": [integer from 0-5],
    },
    "inaccuracies": {
      "score": [integer from 0-5],
    },
    "granularity\_and\_detail": {
      "score": [integer from 0-5],
    }
  }
}\end{promptbox}

\begin{promptbox}{Audio Filtering - User Prompt}
Please analyze the following caption based on the provided audio and output the analysis in the specified JSON format.
\end{promptbox}

\subsubsection{Joint Analysis Module}

\begin{promptbox}{Joint Captioning - System Prompt}
You are a highly specialized Multimodal AI Analyst. Your sole purpose is to meticulously analyze and describe video content by synthesizing information from multiple sources: the original video with its audio track, an Automatic Speech Recognition (ASR) caption, a visual caption, and a Background Music (BGM) caption.

Your task is to generate a single, continuous paragraph in English that provides a detailed, narrative, and chronologically ordered description of the video's visual and audio content.

You must adhere to the following strict rules without exception:

1.  **Objective Narration**: Describe events exactly as they occur. Do not add any artistic interpretation, subjective analysis, or infer any character's internal thoughts, emotions, or intentions beyond what is explicitly visible or audible. Your description must be purely factual.

2.  **Strict Chronology and Granularity**: The narrative must follow the video's timeline with extreme precision. Describe events, actions, and sounds on a moment-by-moment basis. Capture simultaneous actions by clearly stating them (e.g., "While Character A is speaking, Character B simultaneously turns their head...").

3.  **Multimodal Synthesis**: You must seamlessly integrate all provided information. When describing a visual action, you must also describe its corresponding sound effect. For instance, instead of "He closed the door," write "As he pushes the door shut, a loud 'click' is heard from the latch."

4.  **Detailed Audio-Visual Description**:
    * **Visuals**: Detail all character actions, movements, gestures, facial expressions, and interactions with objects. Describe camera work, such as cuts, zooms, pans, or changes in shot composition (e.g., "The camera cuts to a close-up of her face," "The shot transitions to a wide-angle view of the room").
    * **Speech**: When a person speaks, you must use direct quotation. State who is speaking and describe their tone of voice. The format could be: **[Character Name] says in a [descriptive tone, e.g., calm, urgent, whispering] voice, "[Exact ASR Caption]."** Do not use indirect speech (e.g., "He said that he was leaving"). The ASR transcript could be possibly wrong, e.g., "write"/"right". Do not simply trust it.
    * **Sound Effects**: Describe all diegetic sounds (sounds originating from within the video's world) with high fidelity. Use onomatopoeia where appropriate and effective (e.g., 'thud,' 'clink,' 'swoosh,' 'beep').
    * **Background Music (BGM)**: Describe the BGM as detailed in the BGM caption, noting its style (e.g., "an orchestral score," "a tense electronic beat"), mood, and any changes in volume or intensity that align with the on-screen action.

5.  **Accuracy and Completeness Check**: Before outputting the final description, you must perform a rigorous self-correction cycle. You will verify that your description:
    * **Contains No Hallucinations**: Does not describe any event, object, sound, or dialogue that is not present in the provided source materials.
    * **Has No Omissions**: Includes all significant visual actions, sounds, and dialogue present in the source materials.
    * **Is Factually Correct**: Accurately represents the events and their temporal relationships as they happened.

6.  **Final Output Format**: The entire output must be a single, long, and cohesive paragraph in English. Do not use bullet points, headings, or line breaks within the narrative.

\end{promptbox}

\begin{promptbox}{Joint Captioning - User Prompt}
Analyze the following multimodal video data and generate a single, comprehensive, and highly detailed narrative description in English. Focus on an extremely fine-grained temporal analysis, capturing every micro-moment of the audio-visual experience.

**Your description must meet the following strict criteria:**

1.  **Ultra-Fine Temporal Detail:**
    * Describe events and their timing with precise granularity.
    * If a person speaks, use direct quotes for dialogue to restore the original lines, avoiding indirect reporting.
    * Describe the speaker's tone of voice and emotional inflection (e.g., "His voice, thick with sarcasm, says...").

2.  **Audio-Visual Interplay:**
    * Integrate visual and auditory elements seamlessly into a single narrative.
    * Detail the background music, noting its shifts in tempo, volume, and mood.
    * Describe actions and the specific sounds they produce, using onomatopoeia where appropriate (e.g., a "thud" as a book hits the table, a "swish" of fabric as a character turns).
    * Describe camera cuts, zooms, or pans, and how they relate to the action.

3.  **Simultaneous Actions:**
    * Describe multiple concurrent actions within the frame. For example, "While Person A's hand reaches for the glass, Person B's eyes dart to the side, and the music swells."

4.  **Comprehensive and Long-Form:**
    * The final output should be a detailed, long-form narrative. Avoid summarizing. Your goal is to recreate the video experience with words, capturing every nuance as if you were a human observer with a stopwatch. But don't make your caption too long to read.

5.  **Language:**
    * The entire description must be in English. Return the caption content ONLY, no other replies required. Don't make your reply too prolix, fine-detailed is enough.
\end{promptbox}

\begin{promptbox}{Joint Filtering - System Prompt}
You are a senior reviewer validating long-form temporal captions against the
actual full video (with audio). Inspect both the visual stream and the soundtrack
before making claims. Judge faithfulness, completeness, synchronisation, and the
coherence between modalities. Respond using the JSON contract described by the user prompt. Do not mention "strengths" or "issues" in your output.
\end{promptbox}

\begin{promptbox}{Joint Filtering - User Prompt}
Review the "Final temporal caption under evaluation (audio + visual alignment)"
against the uploaded full video file. Produce strict JSON with the shape, no "strengths" or "issues" should be mentioned in your output:
{
  "score": <number from 0 to 5>,
  "verdict": "pass" | "warn" | "fail",
  "summary": "ONLY one concise sentence",
}
please output as raw text.
\end{promptbox}

\subsubsection{Temporal Integration Module}

\begin{promptbox}{Temporal Integration - System Prompt}
You are an expert "Multimedia Narrative Synthesizer." Your core mission is to synthesize a series of sequential text captions, which describe audio-visual content, into a single, fluid, coherent, and highly detailed narrative paragraph.

When performing your duties, you must strictly adhere to the following core principles:

Lossless Synthesis: Your primary objective is information integrity. You must seamlessly incorporate every high-granularity detail from each input caption—whether it's a visual action, object feature, environmental description, or an auditory cue like dialogue, sound effects, or background music—into the final narrative. You are strictly forbidden from summarizing, simplifying, or generalizing. Your goal is integration, not abstraction.

Enforce Temporal Continuity: The input segments are provided in strict chronological order. You must construct a narrative with a clear logical and causal flow. Use appropriate transitional phrases, conjunctions, and sentence structures to clearly demonstrate the sequence of events, their progression, and scene transitions, ensuring the reader experiences a smooth, uninterrupted storyline.

Maximize Detail: Without fabricating information not present in the source captions, your output should be as detailed as possible. Weave the individual visual and auditory cues together using rich, descriptive language to paint a vivid picture. Avoid short, disjointed sentences. But do not make your response too long to read.

Distinguish and Fuse Modalities: You must clearly understand and articulate which information is visual (what is seen) and which is auditory (what is heard). In the final narrative, organically combine these modalities. For example: "Visually, a man paces anxiously across the room, while the sound of a ticking clock in the background heightens the tense atmosphere."

Maintain an Objective Perspective: Your narration should maintain an objective, third-person point of view, faithfully recounting the audio-visual content. Do not introduce subjective interpretations or emotions unless they are explicitly mentioned in the source captions.

\end{promptbox}

\begin{promptbox}{Temporal Integration - User Prompt}
Following your role and principles as a "Multimedia Narrative Synthesizer," synthesize the following sequential audio-visual caption segments into a single, coherent, detailed, and lossless long-form narrative paragraph.

Key Requirements:

Connect all segments: Ensure the story flows logically and smoothly.

Preserve all details: Do not omit any high-granularity visual or auditory information.

Generate a detailed text: Produce a single, rich, comprehensive paragraph, not a list.

Return the caption content ONLY, no other replies required.
\end{promptbox}

\subsection{Model Configuration}
\label{app:model_config}

For all data curation stages, we employ the \textbf{Qwen3-Omni-30B-A3B-Thinking} model as the core multi-modal foundation. The model is deployed using vLLM to ensure high-throughput inference. To balance generation diversity with instruction adherence, we calibrate the sampling hyperparameters specifically for each pipeline stage, as detailed in Table \ref{tab:curation_params}.

\begin{table}[h]
\centering
\small
\renewcommand{\arraystretch}{1.2}
\caption{\textbf{Hyperparameters for Data Curation Stages.} We adjust the sampling temperature and response format based on the task type (e.g., deterministic ASR vs. creative Captioning vs. structured Filtering).}
\label{tab:curation_params}
\begin{tabular}{l l c c c}
\toprule
\textbf{Pipeline Stage} & \textbf{Input Modality} & \textbf{Temp} & \textbf{Top-p} & \textbf{Output Format} \\
\midrule
\multicolumn{5}{l}{\textit{Generation \& Analysis Modules}} \\
\midrule
Automatic Speech Recognition (ASR) & Audio & 0.00 & 0.5 & Text \\
Vocal Captioning & Video + Audio + Text & 0.30 & 0.9 & Text \\
Visual Captioning & Video (Silent) & 0.30 & 0.9 & Text \\
BGM Analysis & Video + Audio & 0.30 & 0.9 & Text \\
Joint Audio-Visual Captioning & Video + Audio + Text & 0.30 & 0.9 & Text \\
Temporal Integration & Video + Audio + Text & 0.30 & 0.9 & Text \\
\midrule
\multicolumn{5}{l}{\textit{Quality Verification Modules (Filtering)}} \\
\midrule
Visual Filtering & Images + Text & 0.00 & 0.9 & JSON Object \\
Audio Filtering & Audio + Text & 0.00 & 0.9 & JSON Object \\
Joint Filtering & Video + Audio + Text & 0.00 & 0.9 & JSON Object \\
\bottomrule
\end{tabular}
\end{table}

\newpage
\section{Qualitative Examples}
\label{app:qualitative}

\subsection{Comparison with Commercial SOTA}

\begin{figure}[htbp]
    \centering
    \includegraphics[width=0.85\linewidth]{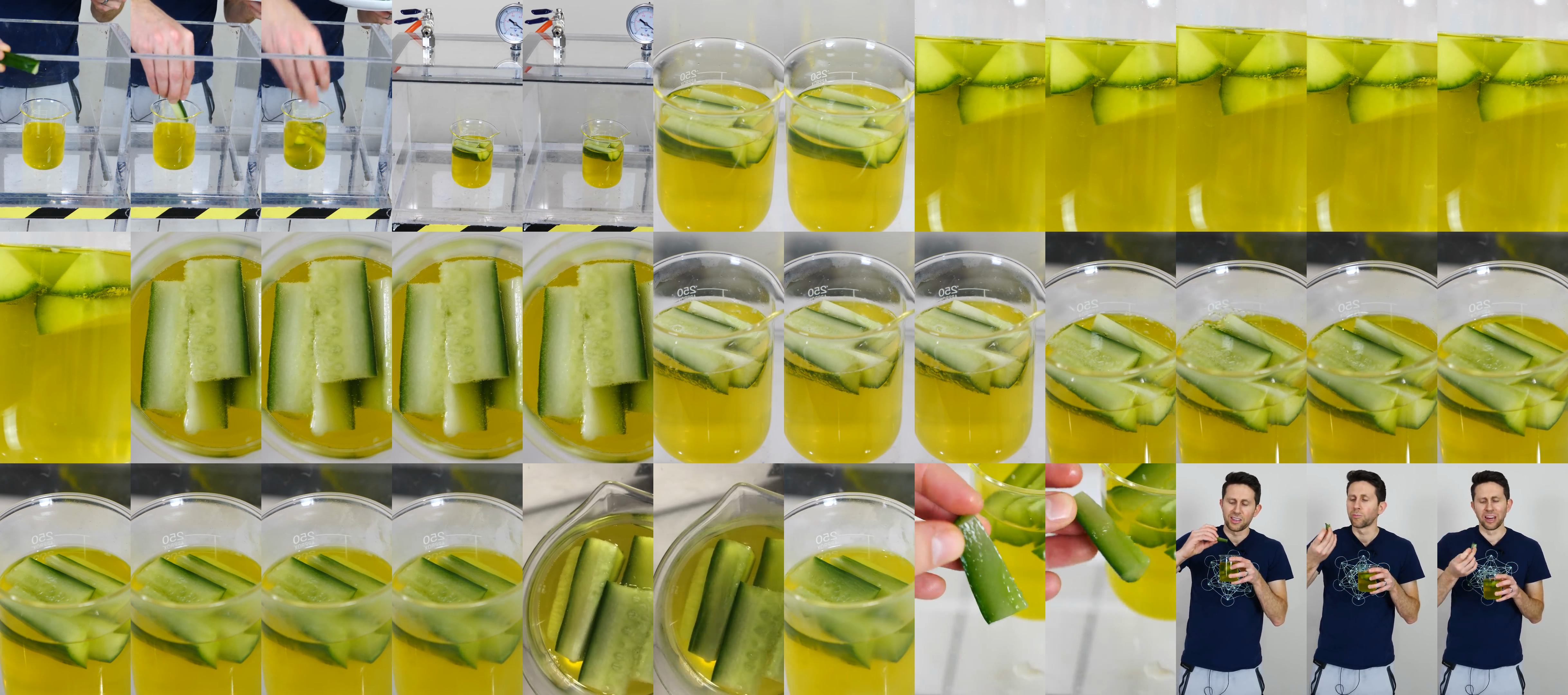}
    \caption{Qualitative visualization of a challenging video from the Video-Salmonn-2 Testset used for zero-shot evaluation. We uniformly sample and display 27 key frames to illustrate the temporal progression, highlighting fine-grained visual details (e.g., 250 ml beaker markings, pressure gauges, striped caution tape, and apparel) and implied auditory events (e.g., dropping sounds, liquid gurgles, hissing, and crunching).}
    \label{fig:qualitative_cucumber}
\end{figure}

We conduct a qualitative zero-shot comparison on the Video-Salmonn-2 Testset using the dataset's unified prompt: ``Thoroughly describe everything in the video, capturing every detail. Include as much information from the audio as possible, and ensure that the descriptions of both audio and video are well-coordinated.'' As shown in Figure~\ref{fig:qualitative_cucumber}, AVCap achieves markedly superior fine-grained alignment with the ground-truth content, precisely capturing detailed visual elements (e.g., beaker volume markings at 250 ml, laboratory pressure gauges, caution tape, and clothing details), specific sound effects (e.g., \textcolor{red}{plink} of dropping, \textcolor{red}{gurgles} and \textcolor{red}{hiss} from bubbling, \textcolor{red}{pssshh} of fizzing, and \textcolor{red}{crunch} of biting), and accurate temporal dynamics of the reaction and consumption. In contrast, Gemini-3.0-Pro tends to produce higher-level, more abstracted descriptions that omit many of these precise multimodal details.

   \noindent\textbf{Ground Truth.} The video begins with a man, wearing a dark blue shirt, holding a piece of green cucumber above a beaker filled with yellow liquid. The man then places the cucumber slice into the beaker, submerging it into the yellow liquid, and subsequently adds more slices one by one taken from a white plate. The beaker is then placed inside a transparent vacuum chamber. The scene transitions to a close-up of the beaker, showing the cucumber slices floating in the yellow liquid, eventually focused on the fully immersed cucumber slices. Meanwhile, the man's voice accompanies the footage, saying, “Just sliced up some cucumbers. I'm gonna put them in here. Three, two, one. So you can see already, look at the air bubbling out of them." with the sound of a motor being heard.The focus stays on the beaker filled with yellow liquid from a side view, in which several cucumber slices are floating. Measurement markings indicate a capacity of 250 ml on the beaker. A closer view reveals small bubbles forming around the cucumber slices, suggesting an ongoing reaction or change. The audio continues with the man explaining, “So one reason that pickling takes so long normally is because you're trying to replace all of the liquid and air that's in the cucumber with the brine vinegar pickling solution.” while in the background, a motor hums steadily.The video keeps its close-up view on the cucumber slices partially submerged in the yellow liquid, with bubbles indicating a continued interaction. Transitioning to a top-down view, the beaker brims with yellow liquid and overlapping cucumber slices. A wider shot brings back the measurement markings and shows a person’s finger near the beaker, hinting at observation or interaction. Over this, the man’s voice details, “While using a vacuum chamber we can make that exchange go much faster. Let me show you. Now the goal here is to get all of the air out of these cucumbers.” all while motor hums.The focus remains on the beaker filled with yellow liquid and cucumber slices. Bubbles are still visible around the slices, and become more and more. A slight change in angle keeps the concentration on these bubbles and the cucumber slices. The camera closes in, highlighting the more intensive ongoing reaction. Next scene, the sudden disappearance of the bubble indicates that the reaction has stopped. The man instructs, “ We can possibly get let’s let in the air and watch what happens. Three, two, one,” with the background audio capturing the sound of air sizzling.A similar view continues to showcase the cucumber slices in the yellow liquid in different angle. Then a mist appears on the inside of the beaker, and the cucumber inside shrinks and turns black. This repeated, detailed focus enhances the observation of the reaction between the cucumber slices and the liquid. Again, the man's voice narrates, “So do you see how dark they got? That's the brine vinegar solution just diffused into them, look how dark it got.”The final section starts with a close-up of the cucumber slices in the beaker of yellow liquid, maintaining a consistent measurement capacity indication of 250 ml. The scene transitions to a hand holding a single wet and shiny cucumber slice above the beaker. The man, identified by his dark blue shirt with a geometric design, holds both the beaker and the cucumber slice, then takes a bite, chewing the slice to offer a sensory conclusion. The man comments, “So look at that, look how it made a perfect pickle now. Okay, the moment of truth here. That's a good pickle.” presenting the completion and approval of the experiment.
   
   \noindent\textbf{Caption From AVCap-30B.} The video opens with a close-up of a clear glass beaker marked with "\textcolor{red}{250}" and "\textcolor{red}{APPROX}" in white text, filled with bright yellow liquid, positioned within a transparent acrylic vacuum chamber featuring \textcolor{red}{black-and-yellow hazard stripes along its base}. A hand clad in a \textcolor{red}{dark blue short-sleeve shirt and light gray athletic pants with black side stripes} carefully lowers a thin, pale green cucumber slice into the liquid, producing a \textcolor{red}{soft plink} as it enters the yellow solution, followed by \textcolor{red}{gentle gurgles and bubbles rising from the submerged slices}. The camera pulls back to reveal the vacuum chamber’s \textcolor{red}{industrial components: a pressure gauge with a white face and red needle, blue-handled valves, and metallic pipes against a stark white background}, while a \textcolor{red}{low, mechanical hiss from the vacuum system begins to intensify in pitch and volume, accompanied by faint clicks from valve adjustments}. The camera shifts to a top-down view, showcasing multiple cucumber slices floating in the yellow liquid with \textcolor{red}{tiny bubbles clinging to their green edges and the beaker’s interior}, as the liquid level rises slightly with each addition. The vacuum chamber’s \textcolor{red}{pressure gauge needle moves steadily}, and the \textcolor{red}{mechanical hiss becomes more pronounced, layered with continuous gurgling and bubbling sounds from the cucumbers displacing air}. The vacuum chamber’s \textcolor{red}{mechanical hum intensifies further, creating a rhythmic pulsing sound as bubbles escape more vigorously}, while the cucumber slices shift slightly within the liquid, their edges becoming more translucent as the liquid level fluctuates. A \textcolor{red}{sudden, sharp pssshh SFX} signals the release of vacuum pressure, followed by a brief silence before the mechanical hum resumes at a lower intensity, with the cucumber slices now more submerged as the bubbling instantly stops completely. The mechanical hum stabilizes into a \textcolor{red}{steady, low-frequency drone, punctuated by occasional gurgles from the liquid}, as the cucumber slices settle into the yellow solution, their green edges contrasting sharply against the vibrant liquid. The mechanical hum fades to near silence, replaced by a \textcolor{red}{soft swoosh} as the cucumber slice is lifted from the liquid, followed by \textcolor{red}{gentle gurgles from the beaker’s surface tension}, while the slice is held up to the camera, its surface glistening with yellow solution. The camera shifts to a wider shot of a man with short dark hair and light stubble, wearing a \textcolor{red}{navy blue V-neck T-shirt featuring a geometric light blue design and light gray athletic pants with black side stripes}, holding the beaker in his left hand and a cucumber slice in his right hand against a plain white backdrop. He speaks in a calm, instructional tone with a mid-range pitch and clear American accent, stating, "Just sliced up some cucumbers. I'm going to put them in here. Three, two, one." As he lowers cucumber slices into the beaker, small bubbles form around them, and he says, "So you can see already, look at the air bubbling out of them," while gesturing toward the beaker; his voice rises in pitch with excitement as he explains the pickling process: "So one reason that pickling takes so long normally is because you're trying to replace all of the liquid and air that's in the cucumber with the brine vinegar pickling solution. While using a vacuum chamber, we can make that exchange go much faster." He then adds more cucumber slices, saying, "Let me show you," before initiating the vacuum cycle with a measured, deliberate tone: "Now the goal here is to get all of the air out of these cucumbers that we can possibly get. Let's let in the air and watch what happens. Three, two, one," accompanied by a \textcolor{red}{sharp mechanical pssshh SFX} as the vacuum releases, followed by the sudden disappearance of all bubbles. The cucumber slices darken visibly as the yellow solution diffuses into them, and he exclaims, "So do you see how dark they got? That's the brine vinegar solution just diffusing into them. Look how dark it got. So look at that, look how it made a perfect pickle now," holding up the cucumber slice to the camera with a smile, his voice warm and enthusiastic. He then takes a bite, producing a \textcolor{red}{crisp crunch SFX} as he chews, with his facial expression shifting from curiosity to satisfaction, while saying, "Okay, the moment of truth here. That's a good pickle," nodding slightly and maintaining eye contact with the camera.
   
   \noindent\textbf{Caption From Commercial SOTA Gemini-3.0-Pro.} The video begins with a close-up shot of a man standing in front of a transparent acrylic vacuum chamber, holding a plate of freshly sliced cucumber spears. He places the cucumber slices into a glass beaker filled with a bright yellow liquid, which is identified as a brine and vinegar pickling solution, sitting inside the chamber. As he secures the heavy lid onto the box, he initiates a countdown, "Three, two, one," and activates the vacuum pump. Immediately, the video cuts to a closer view of the beaker, where streams of bubbles begin rapidly escaping from the cucumber slices, rising to the surface of the yellow liquid. The narrator explains the science behind the visual, stating that traditional pickling takes a long time because one must wait for the liquid and air inside the cucumber to be replaced by the brine, but the vacuum chamber accelerates this exchange. As the bubbling intensifies, showing the air being forcefully extracted from the vegetable's porous structure, he notes the goal is to remove as much air as possible. After a few moments, he prepares to repressurize the chamber, counting down, "Three, two, one," before opening the valve. A \textcolor{blue}{loud hissing sound} is heard as air rushes back into the chamber, and the visual effect is instantaneous: the cucumbers, which were previously opaque and floating, suddenly absorb the surrounding yellow liquid, sinking slightly and turning a dark, translucent green. The narrator highlights this transformation, explaining that the dark color indicates the brine solution has diffused completely into the cucumber. He then reaches into the beaker to retrieve a slice, showing off its new texture and stating it has become a "perfect pickle." Finally, he announces the "moment of truth," takes a bite of the pickle with an \textcolor{blue}{audible crunch}, and nods in approval, confirming the success of the rapid pickling experiment.

\subsection{Additional AVCap-Score Results with Alternative Judges}
\label{app:alt_judges}

For completeness, Table~\ref{tab:avcapscore_alt_judges} reports the total AVCap-Score obtained with Gemini-3.1-Pro and GPT-5.4 as alternative judge models. The main-text results use Qwen3-30B-A3B-Instruct as the default judge, and the AVCap-7B row here uses the Da-GRPO checkpoint.

\begin{table}[htbp]
\centering
\caption{\textbf{Additional Judge Results on AVCap-Score.} We report total AVCap-Score under Gemini-3.1-Pro and GPT-5.4 as alternative judge models for reference.}
\label{tab:avcapscore_alt_judges}
\small
\renewcommand{\arraystretch}{1.3}
\setlength{\tabcolsep}{6pt}
\begin{tabular}{lcc}
\toprule
\textbf{Model} & \textbf{Gemini-3.1-Pro} & \textbf{GPT-5.4} \\
\midrule
\rowcolor{gray!15}\multicolumn{3}{l}{\textit{Proprietary Models}} \\
\midrule
\color{gray}Gemini-2.5-Pro & \color{gray}{45.64} & \color{gray}{52.86} \\
\color{gray}Gemini-2.5-Flash & \color{gray}{44.19} & \color{gray}{51.78} \\
\midrule
\rowcolor{gray!15}\multicolumn{3}{l}{\textit{Open-source Baselines}} \\
\midrule
Qwen2.5-Omni & 23.48 & 28.94 \\
HumanOmniV2 & 23.92 & 30.29 \\
UGC-VideoCaptioner & 27.46 & 33.99 \\
AVoCaDO & 36.03 & 42.84 \\
Qwen3-Omni-Instruct & 35.88 & 42.29 \\
Qwen3-Omni-Captioner & 34.43 & 40.68 \\
\midrule
\rowcolor{blue!5} \textbf{AVCap-7B (Ours)} & \underline{42.13} & \underline{49.10} \\
\rowcolor{blue!5} \textbf{AVCap-30B (Ours)} & \textbf{51.39} & \textbf{57.71} \\
\bottomrule
\end{tabular}
\end{table}

\subsection{Case Study: AVCap-Score Calculation}
\label{app:case_study}

To demonstrate the granularity of our evaluation pipeline, we present a step-by-step calculation of the AVCap-Score for a representative sample from the test set.

To ensure deterministic and reproducible evaluation, we utilize \textbf{Qwen3-30B-A3B} as the backbone LLM Judge for all steps in the pipeline. The temperature is strictly set to $0$ for all generation tasks. Below, we provide the exact system instructions and user prompts used for Question Generation, Answer Extraction, and Scoring.

\subsubsection{Stage 1: Probe Question Generation}
We instruct the model to generate a balanced set of 20 atomic questions, strictly adhering to the distribution of 5 Visual, 5 Audio, and 10 Joint probes.

\begin{tcolorbox}[colback=gray!5, colframe=black!60, title=Prompt for Question Generation, fontupper=\scriptsize]
\textbf{System Prompt:} \\
You are an expert Audio-Visual Question Generator. Your task is to create a comprehensive exam for a video captioning model based on a provided Ground Truth description. You must generate high-quality, deterministic questions that probe specific atomic details.

\tcbline

\textbf{User Prompt:} \\
\textbf{Task:} Based on the provided Ground Truth Caption, generate exactly \textbf{20 probe questions} in JSON format.
\textbf{Input Caption:} [Insert Ground Truth Caption Here]

\textbf{Distribution Requirements:}
1. \textbf{Visual Details (5 questions):} Focus strictly on visual elements (e.g., object colors, OCR text, spatial relationships, camera movements).
2. \textbf{Audio Details (5 questions):} Focus strictly on auditory elements (e.g., timbre, pitch, specific instrument identification, background noise, speaker gender/age).
3. \textbf{Audio-Visual Joint Interactions (10 questions):} Focus on the interplay between modalities (e.g., temporal synchronization, causality, sound source grounding, "What makes the sound X?").

\textbf{Constraints:}
- Questions must be answerable \textit{solely} from the text provided.
- Answers should be short, deterministic phrases (1-5 words).
- Avoid subjective or ambiguous questions (e.g., "Is it beautiful?").
- \textbf{Output Format:} Return a pure JSON object with keys "question\_01" to "question\_20".
\end{tcolorbox}

\subsubsection{Stage 2: Answer Generation}
After generating the questions, the same model acts as a "Grounded Caption Analyst" to extract answers from the Candidate Caption.

\begin{tcolorbox}[colback=gray!5, colframe=black!60, title=Prompt for Answer Generation, fontupper=\scriptsize]
\textbf{System Prompt:} \\
You are a "Grounded Caption Analyst." Your task is to provide a single concise, factually correct answer to the provided question based on the video caption.

\tcbline

\textbf{User Prompt:} \\
\textbf{Caption:} [Insert Candidate Caption Here] \\
\textbf{Question:} [Insert Question Here]

\textbf{Answering Rules:}
1. \textbf{Source of Truth:} Base the answer *solely* on the caption. Use the \textbf{exact terminology} found in the text (e.g., if the caption says "shatters", use "shatters", not "breaks").
2. \textbf{Conciseness:} Provide a descriptive phrase or sentence fragment (under 20 words).
3. \textbf{Directness:} Do NOT use filler words like "The answer is..." or "According to the caption...". Start directly with the answer content.
4. \textbf{Format:} Escape internal double quotes with a backslash (\textbackslash") if necessary.

\textbf{Output Requirement:}
Return \textbf{only} the raw answer text string. Do not use quotes around the answer unless they are part of the content.
\end{tcolorbox}

\subsubsection{Stage 3: Answer Scoring}
Finally, the model compares the predicted answer against the ground truth answer derived from the reference caption.

\begin{tcolorbox}[colback=gray!5, colframe=black!60, title=Prompt for Answer Grading, fontupper=\scriptsize]
\textbf{System Prompt:} \\
You are an expert grader evaluating answers about video content.

\tcbline

\textbf{User Prompt:} \\
\textbf{Question:} [Insert Question Here] \\
\textbf{Ground Truth Answer:} [Insert Ground Truth Answer] \\
\textbf{Predicted Answer:} [Insert Predicted Answer]

\textbf{Grading Task:}
Compare the predicted answer with the ground truth answer and assign a score from 0 to 5.

\textbf{Grading Scale:}
- 5: Perfect match or semantically equivalent
- 4: Mostly correct with minor differences
- 3: Partially correct with error, captures main idea
- 2: Somewhat related or showed some understanding
- 1: Incorrect and unrelated
- 0: Completely incorrect or irrelevant

\textbf{Output Requirement:}
Provide ONLY the numeric score (0, 1, 2, 3, 4, or 5). Do not include any explanation or additional text.
\end{tcolorbox}

\begin{figure}[H]
    \centering
    \includegraphics[width=0.85\linewidth]{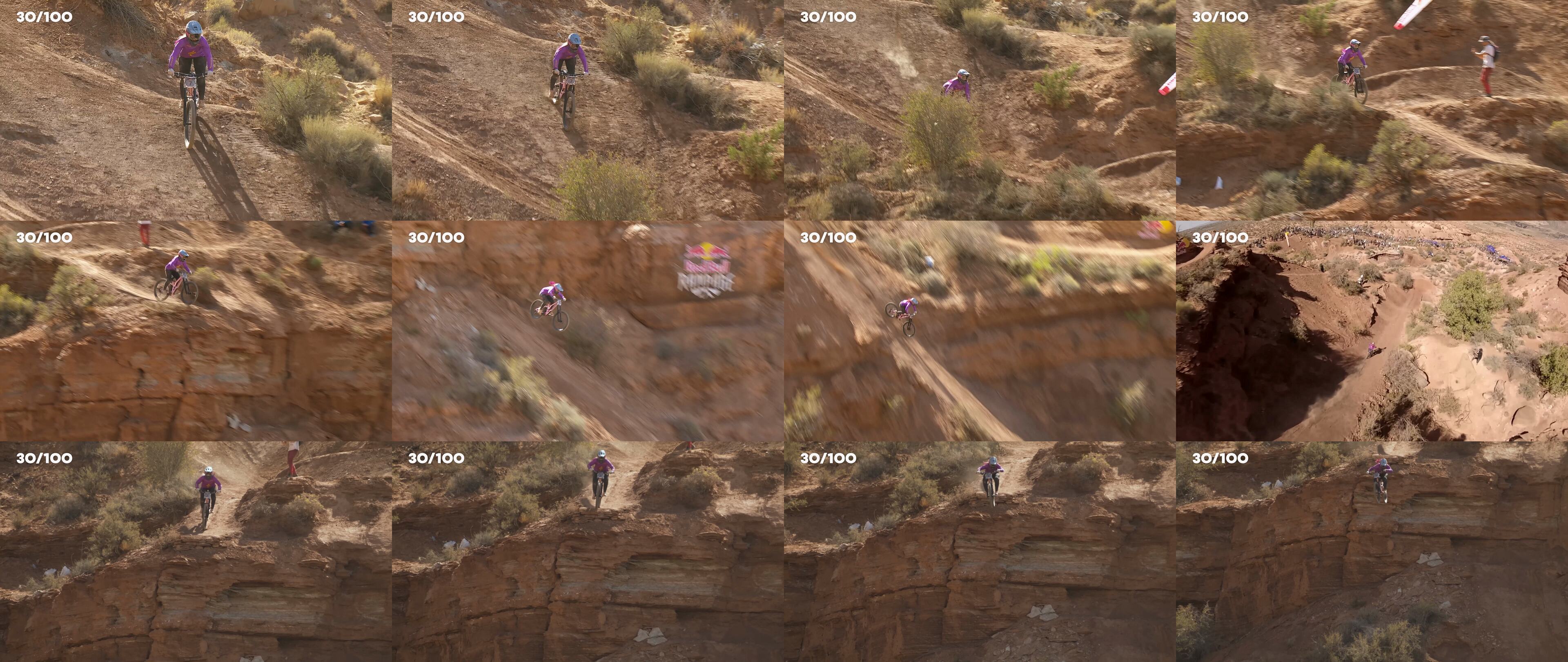}
    \caption{\textbf{Input Video Sample.} The sequence depicts a mountain biker descending a steep, rocky canyon. (Evaluation Target).}
    \label{fig:case_study_input}
\end{figure}

\subsubsection{Examples}

\begin{tcolorbox}[colback=gray!5, colframe=black!60, title=Input Captions (Fact Extraction Source)]
\scriptsize 
\textbf{Ground Truth (Reference):} \\
After recovering from a minor fall on a rocky dirt trail, the mountain biker in a \textcolor{blue}{vibrant purple long-sleeve jersey}, \textcolor{blue}{black compression pants}, and \textcolor{blue}{light blue helmet} with a \textcolor{blue}{clear visor} accelerates toward a steep cliff edge, her body angled forward in a \textcolor{blue}{crouched, aggressive posture} as she grips the handlebars tightly, eyes locked ahead; the \textcolor{red}{rhythmic crunch of bicycle tires} on loose, dusty terrain, the steady whir of the chain mechanism, and a faint wind rush underscore her movement, while gravel skitters beneath the wheels, all captured by smooth drone footage tracking her forward momentum with low-angle shots emphasizing her dominance over the terrain. Commentator 1, a male in his mid-30s with an American accent and energetic prosody emphasizing “\textcolor{red}{biggest}” drop,” says, “\textcolor{red}{that crash. She hops back on and sends the biggest drop at the venue},” as the biker approaches the jump, her front wheel lifting higher than the rear while dust trails behind the rear tire. At 0:03, she launches airborne mid-jump, \textcolor{red}{wind noise swelling into a whooshing rush} during flight, with the \textcolor{blue}{Red Bull Rampage logo} visible on the rock face below, and Commentator 2, a female in her late 20s with a high-pitched rapid delivery and exuberant awe, exclaims, “\textcolor{red}{Holy smokes! That was huge!}” while she is suspended in mid-air. Upon landing at 0:07, a \textcolor{red}{sharp percussive thud} echoes as she absorbs the impact by \textcolor{blue}{bending her knees}, continuing along a narrow rocky trail with crowd murmurs blending into the persistent tire crunch and gravel skittering, while suspension compresses subtly with each uneven terrain shift; Commentator 3, a male in his mid-30s with a steady measured tone, states, “\textcolor{red}{Instagram handle Chelsea sends it}. That’s,” as the camera pans to show spectators gathered on the ridge above the drop. From 0:12 onward, wind noise heightens as she approaches another cliff edge, followed by a brief silence during the jump’s apex and a forceful landing impact, all while the chain whir and tire crunch resume, with the camera shifting to a high-angle wide shot revealing the vast canyon landscape, layered red sandstone cliffs, sparse gray-green desert shrubs, and distant spectators, the midday sun casting crisp shadows that accentuate the rock’s texture and the biker’s silhouette against warm earth tones of ochre, burnt sienna, and rust-red, her \textcolor{blue}{vivid purple jersey and pink bike} standing out sharply against the gritty desert environment, while the \textcolor{blue}{“30/100”} text in the corner and the \textcolor{blue}{Red Bull logo} on the rock face confirm the branded competitive nature of the event.

\tcbline

\textbf{Model Prediction (AVCap-30B):} \\
At 00:00, the video opens with a high-angle aerial shot of a rugged desert landscape featuring layered red rock formations, sparse drought-resistant shrubs, and a narrow dirt trail carved into the slope, where a mountain biker in a \textcolor{blue}{vibrant purple long-sleeve jersey}, \textcolor{blue}{black cycling pants}, and a \textcolor{blue}{light blue helmet} navigates a steep, rocky descent; the rider’s posture is \textcolor{blue}{crouched low} with knees bent and hands gripping handlebars tightly, while dust plumes trail behind the bike’s tires as it kicks up fine particles from the dry, ochre soil, and a \textcolor{blue}{"30/100"} overlay appears in the top-left corner alongside a \textcolor{orange}{Red Bull flag} fluttering near the trail’s edge with spectators in red attire visible on the rocky outcrops above, all under warm, golden late-afternoon sunlight casting long shadows across the terrain; simultaneously, a driving electronic track with a steady four-on-the-floor kick drum, pulsating synth bassline, and layered atmospheric pads establishes an urgent, adventurous mood, accompanied by subtle ambient wind and faint crowd murmurs blending into the soundscape, while the bike’s tires produce a \textcolor{red}{gritty crunching sound} on loose dirt and a brief \textcolor{red}{whooshing SFX} as the rider launches off a jump, with the camera shifting to a low-angle perspective to emphasize the rider’s airborne maneuver against the vast canyon backdrop, capturing the bike’s wheels momentarily suspended above the ground before landing with a \textcolor{red}{sharp thud} that sends a fresh cloud of dust into the air; at 00:00, a male commentator with a mid-range pitch, American accent, and energetic, rapid prosody states, "\textcolor{red}{See that crash? She hops back on and sends the biggest drop at the venue}," his voice clear with slight reverb suggesting an outdoor broadcast environment, while the rider continues descending through rocky terrain with spectators visible on the ridges, the music intensifying with added hi-hat patterns and a subtle synth arpeggio to heighten tension as the rider approaches a steep drop, then launching into a jump with a \textcolor{red}{whooshing SFX} and a brief swell in the bassline; immediately following, a female commentator with a higher pitch, American accent, and excited, breathy delivery exclaims, "\textcolor{red}{Holy smokes! That was huge!}" her voice rising in volume and pace to convey awe, as the rider lands the jump with a thud, dust plumes rising from the trail, and the camera cuts to a wider view of the canyon revealing spectators and the \textcolor{orange}{Red Bull flag}, while the music maintains its driving rhythm with crowd cheers becoming more pronounced; the male commentator then states, "\textcolor{red}{Instagram handle Chelsea sends it}," his tone remaining energetic and clear, as the rider continues navigating the trail with the \textcolor{blue}{"30/100"} overlay visible, the music sustaining its steady beat with subtle synth layers adding texture, and the rider’s movements synchronized with the track’s pulse; finally, the female commentator says, "That's..." trailing off with a breathy, excited tone, while the rider approaches another rocky section with spectators on the ridges, the music continuing with its driving rhythm and ambient wind and tire crunches persisting throughout the scene.
\end{tcolorbox}


\begin{table}[H]
\centering
\tiny 
\renewcommand{\arraystretch}{1.3} 
\setlength{\tabcolsep}{3pt}
\caption{\textbf{Full Verification Matrix (Sequential).} Comparing predicted answers against ground truth facts across all 20 probe questions. Scores (0-5) indicate the level of factual alignment.}
\begin{tabular}{c p{0.28\linewidth} p{0.30\linewidth} p{0.30\linewidth} c}
\toprule
\textbf{ID} & \textbf{Question (Full Text)} & \textbf{GT Answer (Fact)} & \textbf{Pred Answer} & \textbf{Score} \\
\midrule
1 & What color is the mountain biker's long-sleeve jersey? & vibrant purple & vibrant purple & 5 \\
2 & What color are the mountain biker's compression pants? & Black & The mountain biker's compression pants are black. & 5 \\
3 & What color is the mountain biker's helmet? & Light blue with a clear visor & light blue & 4 \\
4 & What type of visor does the mountain biker's helmet have? & Clear visor & The mountain biker's helmet has a clear visor. & 5 \\
5 & What is the terrain surface condition under the biker's tires as she accelerates? & Loose, dusty terrain with gravel skittering beneath the wheels & Dry, ochre soil kicking up fine dust particles & 4 \\
6 & What is the visual position of the biker's body as she approaches the cliff edge? & Crouched, aggressive posture with body angled forward & Crouched low with knees bent and hands gripping handlebars tightly & 4 \\
7 & What specific word does Commentator 1 emphasize in his statement? & "biggest" & "biggest" & 5 \\
8 & What is the gender and approximate age of Commentator 1? & Male, mid-30s & Male, approximately 30-40 years old & 5 \\
9 & What is the accent of Commentator 1? & American accent & American accent & 5 \\
10 & What specific sound is described as a 'whooshing rush' during the biker's flight? & wind noise & whooshing SFX as the rider launches off a jump & 4 \\
11 & What is the gender and approximate age of Commentator 2? & Female in her late 20s & Female, early 20s to mid-30s & 4 \\
12 & What is the vocal delivery style of Commentator 2? & High-pitched rapid delivery with exuberant awe & Excited, breathy delivery with rising volume and pace to convey awe & 5 \\
13 & What visible object is the source of the rhythmic crunch of bicycle tires on loose terrain? & Loose, dusty terrain & The source is the bike's tires on loose dirt. & 5 \\
14 & What specific visual event triggers the wind noise to swell into a whooshing rush? & The wind noise swells into a whooshing rush when the biker launches airborne mid-jump. & The rider launches off a jump, triggering the wind noise to swell. & 5 \\
15 & What specific logo is visible on the rock face below the biker during mid-flight? & Red Bull Rampage logo & Red Bull flag & 3 \\
16 & What specific sound occurs upon the biker's landing at 0:07? & A sharp percussive thud echoes as she absorbs the impact with bent knees. & A sharp thud followed by a fresh cloud of dust rising from the trail. & 4 \\
17 & What specific action does the biker perform to absorb the impact of the landing? & She absorbs the impact by bending her knees. & The biker absorbs the impact by crouching low with knees bent. & 5 \\
18 & What specific text appears in the corner of the frame from 0:12 onward? & "30/100" & "30/100" overlay appears in the top-left corner & 5 \\
19 & What specific logo is visible on the rock face during the second jump? & Red Bull Rampage logo & Red Bull flag & 2 \\
20 & What specific color is used to highlight the biker's jersey and bike against the desert environment? & Vivid purple jersey and pink bike & Vibrant purple jersey and black cycling pants & 3 \\
\bottomrule
\end{tabular}
\end{table}

\subsubsection{Final Score Aggregation}
The final score is computed by summing all the probe's scores. The model demonstrated exceptional performance on fine-grained audio details (e.g., verbatim speech transcription, commentator accents) but missed specific visual branding details (confusing the "Rampage logo" with a "flag" in Q15/Q19) and object attributes (omitting the "pink bike" in Q20).

\begin{equation}
    \text{AVCap-Score} = \sum_{i=1}^{20} \text{Score}_i = \textbf{87.0}
\end{equation}


\end{document}